\documentclass{article} % For LaTeX2e

\usepackage{iclr2026-conference,times}
\usepackage{amsmath,amsfonts,bm}

\def\eqref#1{equation~\ref{#1}}
\def\1{\bm{1}}

\DeclareMathAlphabet{\mathsfit}{\encodingdefault}{\sfdefault}{m}{sl}
\SetMathAlphabet{\mathsfit}{bold}{\encodingdefault}{\sfdefault}{bx}{n}

\usepackage{hyperref}
\usepackage{url}
\usepackage{graphicx}
\usepackage{xspace}
\usepackage{makecell}
\usepackage{array}
\usepackage{colortbl}
\usepackage{multirow}
\usepackage[normalem]{ulem}
\usepackage{subcaption}
\usepackage{tabularx}
\usepackage{color}
\usepackage{pifont}
\usepackage[utf8]{inputenc}
\usepackage[T1]{fontenc}
\usepackage{CJKutf8}
\usepackage{enumitem}
\usepackage{diagbox}
\usepackage{listings}
\usepackage[flushleft]{threeparttable}
\usepackage{booktabs}
\usepackage{wrapfig}
\usepackage{algorithmicx,algorithm}
\usepackage{xcolor}
\definecolor{redorange}{rgb}{1, 0.5, 0}
\definecolor{darkgreen}{rgb}{0.0, 0.5, 0.0}
\definecolor{darkred}{rgb}{0.7, 0.0, 0.0}
\usepackage{pifont}
\usepackage{tikz}
\usepackage{bm}
\newcommand{\with}{\textcolor{darkgreen}{\textbf{\checkmark}}}  % 绿色对勾
\newcommand{\without}{\textcolor{darkred}{\ding{55}}} 

\usepackage{array}      % 提供更灵活的列格式控制
\usepackage{booktabs}   % 提供更好看的表格线 (toprule, midrule, bottomrule)

\newcommand{\rot}[1]{\rotatebox[origin=c]{90}{#1}}

\newcolumntype{C}[1]{>{\centering\arraybackslash}p{#1}}

\newcommand{\icon}{\raisebox{-4.1pt}{\includegraphics[width=1.2em]{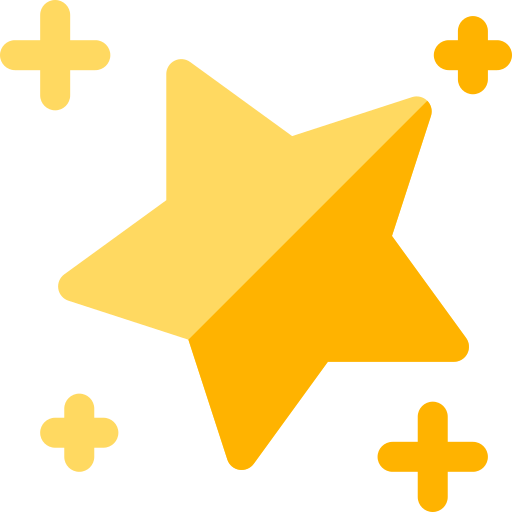}}\xspace}

\title{\icon STAR: boosting time series foundation models for anomaly detection  through STate-aware AdapteR}

\author{Hanyin Cheng$^{1}$, Ruitong Zhang$^{1}$, Yuning Lu$^{1}$, Peng Chen$^{2}$, Meng Wang$^{2}$, Yang Shu$^{1}$, \\\textbf{Bin Yang$^{1}$, Chenjuan Guo$^{1}\thanks{Corresponding author}$}\\
$^1$East China Normal University
\\$^2$2012 APPLab, Huawei\\
\texttt{\{hycheng,rtzhang,ynlu\}@stu.ecnu.edu.cn}, \\
\texttt{\{chenpeng192,wangmeng71\}@huawei.com}, \\
\texttt{\{jlhu,byang,cjguo\}@dase.ecnu.edu.cn}, \\
}

\iclrfinalcopy % Uncomment for camera-ready version, but NOT for submission.
\begin{document}

\maketitle

\begin{abstract}
While Time Series Foundation Models (TSFMs) have demonstrated remarkable success in Multivariate Time Series Anomaly Detection (MTSAD), however, in real-world industrial scenarios, many time series comprise not only \textit{numerical variables} such as temperature and flow, but also numerous discrete \textit{state variables} that describe the system status, such as valve on/off or day of the week.
Existing TSFMs often overlook the distinct categorical nature of state variables and their critical role as conditions, typically treating them uniformly with numerical variables. This inappropriate modeling approach prevents the model from fully leveraging state information and even leads to a significant degradation in detection performance after state variables are integrated.
To address this critical limitation, this paper proposes a novel \uline{\textbf{ST}}ate-aware \uline{\textbf{A}}dapte\uline{\textbf{R}} (STAR). STAR is a plug-and-play module designed to enhance the capability of TSFMs in modeling and leveraging state variables during the fine-tuning stage. Specifically, STAR comprises three core components: (1) We design an \textit{Identity-guided State Encoder}, which effectively captures the complex categorical semantics of state variables through a learnable \textit{State Memory}. (2) We propose a \textit{Conditional Bottleneck Adapter}, which dynamically generates low-rank adaptation parameters conditioned on the current state, thereby flexibly injecting the influence of state variables into the backbone model. (3) We also introduce a \textit{Numeral-State Matching} module to more effectively detect anomalies inherent to the state variables themselves. Extensive experiments conducted on real-world datasets demonstrate that STAR can improve the performance of existing TSFMs on MTSAD.

\end{abstract}
% \begin{center}
% \textbf{Resources:} \href{https://anonymous.4open.science/r/Aurora-40AB}{https://anonymous.4open.science/r/Aurora-40AB}.
% \end{center}

\section{Introduction}

% 近年来，时间序列基础模型Time Series Foundation Models, TSFMs已成为时间序列预测领域一个极具变革性的研究方向。 受到自然语言处理领域大型语言模型成功的启发，研究者们致力于构建能够理解和处理各类时间序列数据的通用模型。
In recent years, research into Time Series Foundation Models (TSFMs) has emerged as a significant area of interest.~\citep{ timemoe, liu2025sundial, tinyttm}. 
% Inspired by the success of Large Language Models in the field of Natural Language Processing, 
Researchers are committed to building general-purpose models capable of understanding and processing a diverse range of time series tasks~\citep{timer,AimTS,units}. Benefiting from large-scale pre-training, TSFMs can deliver competitive performance without requiring extensive, task-specific fine-tuning.
Multivariate Time Series Anomaly Detection (MTSAD), which focuses on identifying abnormal data in multivariate time series, is one of the key downstream tasks targeted by TSFMs~\citep{shentu2024towards,goswami2024moment,wang2025lightgts}. It is applied widely in numerous critical domains, including financial fraud detection, medical disease identification, and cybersecurity threat detection~\citep{wen2022robust,yang2023sgdp,kieu2018outlier}.
% 尽管TSFMs在MTSAD领域中已经取得了令人印象深刻的效果，但是在真实的复杂工业场景中，时间序列的变量具有更复杂的表现形式，除了常见的数值类型的变量（温度，水流量，速度等），还有一些描述系统特定状态的变量（如阀门的开关与否，星期，机器的位置状态等），与数值变量不同的是这类变量在组织形式上呈现出离散的数据形态，每个值代表该状态的具体类别。

\begin{figure*}[!tbp]
% \vspace{-3mm}
    \centering
\includegraphics[width=1\linewidth]{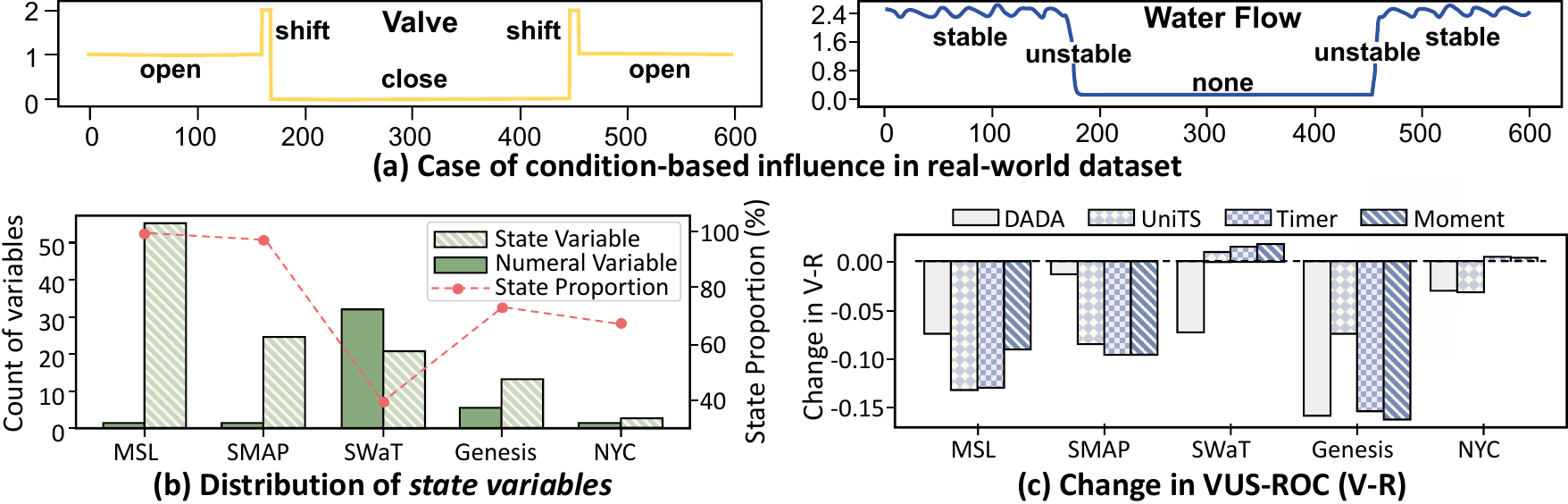}
     \caption{(a) Case of State variables (Valve) and  numeral variables (Water Flow) in SWaT. (b) State variables are prevalent in real-world datasets and frequently constitute a substantial proportion. (c) Impact of incorporating state variables on detection performance.
}

\label{fig: intro}
\vspace{-7mm}
\end{figure*}

Although TSFMs have achieved impressive performance in MTSAD, real-world industrial scenarios present more complex data manifestations. In these settings, variables exhibit more intricate forms. Beyond common \textit{numerical variables} such as temperature, water flow, and velocity, \textit{state variables} also well exist that describe specific system states, such as the open/closed state of a valve~\citep{swat}, the day of the week~\citep{nyc}, or the positional state of a machine~\citep{Genesis}. Unlike their numerical counterparts, these state variables are often organized as discrete data, where each value represents a specific category of that state. 
Moreover, they provide essential context about the system's operational state and directly influence the numerical variables' magnitude or temporal patterns. We refer to such influence as \textit{condition-based influence}. For example, as shown in Figure~\ref{fig: intro}a, the valve's state (open, closed, or shifting) is a precondition for water flow and directly dictates the flow pattern (stable, unstable, or none).
Figure~\ref{fig: intro}b illustrates that state variables are widely prevalent in common datasets. 
% 这些状态对于异常检测十分重要，如The NYC taxi dataset  has two state variables: time of day and day of the week. The same number of passengers carries different implications depending on whether it is daytime or nighttime, a weekday or a weekend. These states impose conditions that affect passenger volume.
% The NYC taxi dataset~\citep{nyc}, for example, has two state variables: time of day and day of the week. The same number of passengers carries different implications depending on whether it is daytime or nighttime, a weekday or a weekend. These states impose conditions that affect passenger volume.
However, due to either a lack of pre-training on state variables or uniform modeling for both numerical and state variables, \textbf{TSFMs often struggle to handle state variables}.  As shown in Figure~\ref{fig: intro}c, when these state variables are incorporated into representative TSFMs through unified modeling, the detection performance fails to improve significantly or suffers a substantial decline.

To address this limitation, we aim to develop a plug-and-play method that enables TSFMs to leverage state variables better, thereby enhancing detection performance and facilitating adaptation to a wider range of industrial scenarios. However, this faces the following challenges:

\textbf{First, how to extract meaningful embeddings from state variables?} 
% 不同的状态变量的相同数值可能具有相似语义信息或异质语义信息，
% 状态变量具有复杂的信息，不同的状态变量可能具有相似语义信息或异质语义信息，总体呈现出聚集特性。
State variables possess complex semantics.
For instance, within the SWaT~\citep{swat}, several state variables are primarily grouped into three clusters: switches of valves,  pumps, and mercury lamps. 
% 同一组内具有相似的语义信息，对系统有着相似的物理作用。不同组间具有heterogeneous的语义信息，对系统有着不同的作用方式。
Within a given group, state variables share similar semantics and exert similar physical influence on the system. Conversely, state variables across different groups are semantically heterogeneous, manifesting in disparate modes of influence.
% where each state variable may be semantically similar to some variables but heterogeneous from others. Collectively, they tend to cluster.
% For instance, within the SWaT~\citep{swat}, state variables are primarily grouped into three clusters: valve switches,  pump switches, and mercury lamp switches.
% Similarly, in the Genesis dataset from the machinery domain, state variables form distinct subgroups representing positional states, slide rail positions, and sensor statuses, among others. 
% 如在Water treatment领域的SWaT数据集中，状态变量可主要分为阀门开关、泵开关、汞灯开关，在Machinery领域的Genesis数据集中有一部分变量代表位置状态，一部分变量代码滑轨位置，一部分变量代表传感器状态等。
% 此外，状态变量还表现出一定的时序模式，呈现出状态的类别和时序波动两方面特征。
Furthermore, state variables also exhibit distinct temporal patterns, such as the stable periodic patterns of variables representing the day of the week~\citep{nyc}.
% which in turn give rise to a two-fold characteristic:  categorical information and temporal fluctuations.
% 在目前的Founadtion models 中， 使用在数值变量预训练好的基于线性层的embdding层中完成嵌入，尽管这种方法可以一定程度上提取状态变量的时序特征，但是却不能提取复杂且丰富的状态类别信息。
In existing foundation models, state variables are typically embedded using linear layers pre-trained on numerical data. While this approach can partially capture the temporal features, it fails to adequately extract the rich and complex semantics inherent in their states.

\textbf{Second, how to model the condition-based influence of state variables?} 
Different from the correlation among variables~\citep{qiu2025duet,wucatch}, the state variables act as preconditions that directly determine the magnitude or temporal patterns of the numerical variables. Therefore, for anomaly detection, it is essential to consider the condition-based influence when processing numerical variables.  However, this asymmetric influence between state and numerical variables renders the modeling of condition-based influence a non-trivial task. Some TSFMs, such as UniTS~\citep{units}, model interactions among all variables in a unified manner, this approach erroneously treats them as equivalent, thereby failing to model the condition-based influence effectively.
% 然而，正是state variables 和 Numeral variables之间这种不平等的，非对称的关系，使得condition-based influence难以被建模。

% % 状态变量代表着某一时刻，系统的某方面具体的特征类别。它不仅影响着我们对于数值变量的理解方式，更在数值信息交互以及异常信息鉴别等方面发挥着重要作用。这种作用机制难以附加给预训练好的基础模型
% % 状态变量作为系统运行的大背景，是数值变量产生时的先决条件。
% By establishing the broad context for the system's operation, state variables serve as a precondition for the generation of numerical variables. 
% % 相同的数值在不同的状态中也代表了不同的物理意义。在进行异常检测时，更是要结合当前系统的状态来判断数值是否合理。
% The same numerical value can signify different physical phenomena in different system states. Therefore, for anomaly detection, it is essential to consider the current system state when processing numerical variables.
% However, this condition-based influence is difficult to integrate into pre-trained TSFMs effectively. While some TSFMs, such as UNITS~\citep{units}, model interactions between \textit{state} and numerical variables in a unified manner, this approach erroneously treats them as equivalent, thereby ignoring the fundamental role of state variables as preconditions.
% % this approach merely enhances the numerical information interactions and fails to genuinely model the conditional operational mechanism of state variables.
% % 尽管一些基础模型，将状态与数值变量统一进行变量交互，但这种方式只是简单的增强了数值信息交互，而无法真正的在整个处理过程中建模状态变量的conditional operational mechanism

To close these gaps, we propose \textbf{STAR}, a novel \textbf{ST}ate-aware \textbf{A}dapte\textbf{R} for TSFMs in the MTSAD to better model state variables during the fine-tuning phase. 
% 首先，为了更好地提取状态变量的features，我们设计了一个可以同时提取状态变量的类别特征以及时序模式的State Extractor，该模块包含了identity-guided State Encoder 通过建立一个小规模的State Book，并在Variable Identity 和 state Identity的指导下，自适应地从中获取一个表征组合，从而有效地建模了状态变量间复杂的同质性与异质性关系。
\textbf{First}, to better extract embeddings from state variables, we designed a \textit{State Extractor} capable of simultaneously capturing both the state information and temporal patterns of state variables. This module incorporates an \textit{Identity-guided State Encoder} which establishes a small-scale State Memory and retrieves a composite memory's representation to encode state information guided by both Variable Identity and State Identity. This method effectively modeling the complex semantics of state variables.
% First, to better extract features from state variables, we designed a State Extractor that incorporates an Identity-guided State Encoder. This encoder establishes a small-scale State Memory and retrieves a composite representation from it, guided by both Variable and State Identities. This approach effectively models the complex semantics in state variables. Furthermore, the State Extractor is capable of simultaneously capturing both categorical information and temporal patterns of state variables.
% 其次，我们提出了一个 Stat-aware boottleneck lora, 通过以状态变量为条件生成动态的lora参数，该模块可以更好的为模型纳入状态变量的影响，同时，我们为该模块设计了动态的瓶颈层，来更灵活的调节模块的信息密度。
\textbf{Second}, we propose a \textit{Conditional Bottleneck Adapter} to effectively modeling the condition-based influence. The adapter dynamically generates adaptation parameters and bottleneck size based on state variables to modulate the TSFM’s parameters, thereby directly influencing how the TSFMs process numerical variables.
%   We also design a dynamic bottleneck mechanism to regulate the information density flexibly.
% 在以上两个模块的基础之上，我们已经可以执行状态感知的微调，此外，为了更加有效地检测状态变量中包含的异常，我们进一步设计了Numeral - State Matching 模块。
Building upon the aforementioned modules, we can already capture meaningful state information and process numerical variables based on state.
Furthermore, we have developed a \textit{Numeral-State Matching} module to more effectively detect anomalies contained within the state variables.

Our contributions are summarized as follows:

 \begin{itemize}[left=0.3cm]

\item  We design a universal State-aware Adapter that equips TSFMs with a more effective mechanism for handling state variables for the MTSAD.

\item   We propose a novel \textit{Identity-guided State Encoder} in STAR to effectively model the complex homogeneous and heterogeneous semantics of state variables.

\item  We propose a novel \textit{Conditional Bottleneck Adapter} in STAR, which can flexibly incorporate the condition-based influence of state variables in the TSFMs.

\item We conducted extensive experiments on real-world datasets. The results show that STAR effectively improves the performance of TSFMs in MTSAD. 

\end{itemize}

\section{Related works}
\subsection{Multivariate time series anomaly detection}
Classic methods for MTSAD can be classified into non-learning~\citep{breunig2000lof,Li_2024_BMVC}, machine learning~\citep{liu2008isolation,ramaswamy2000efficient}, and deep learning~\citep{wucatch, zhong2025simad, liu2024time, liu2024elephant}. In particular, methods based on deep learning have seen significant success in recent years. They can be classified into forecasting-based~\citep{deng2021graph}, reconstruction-based~\citep{wu2022timesnet,luo2024moderntcn,tuli2022tranad} and contrastive-based methods~\citep{xu2021anomaly, yang2023dcdetector,guo2023contranorm}. 
Although some of these models consider variable interactions within a unified modeling framework~\citep{wucatch,xie2025multivariate}, or employ specialized modeling for discrete variables~\citep{li2024hybrid,chen2025unified}, they still treat state variables as mere discrete numerical values, thereby overlooking the condition-based influence of state variables.
% 尽管其中有一些模型在统一建模的基础上考虑了变量交互，或是针对离散变量专门的建模，但是他们仍然将状态变量视作离散的数值，忽略了状态变量本身对于系统condition-based 的影响
% li2024hybrid,chen2025unified

\subsection{Time series foundation models for anomaly detection}
Researchers are actively exploring Foundation Models for various time series analysis tasks, including forecasting~\citep{wang2025lightgts,liu2025sundial,wang2024rose,timesfm}, classification~\citep{AimTS}, and anomaly detection~\citep{shentu2024towards}. 
% 其中，对于异常检测来说，可以分为两类：1）task-通用模型 这类模型通常通过在广泛数据上进行预训练，具备出色的表征学习能力，通过重构进行异常检测，2）task-专属模型。这类模型针对于异常检测任务进行模型设计，如DADA为了更好地适应异常数据，使用多个不同尺寸的瓶颈层进行解码。
For anomaly detection, models can be classified into two primary categories: 1) \textbf{Task-general model}: These models are typically pre-trained on extensive data, endowing them with excellent representation learning capabilities~\citep{units,goswami2024moment,timer}. 2) \textbf{Task-specific model}: These models have architectures specifically designed for the anomaly detection task. For instance, to better adapt to anomalous data, DADA~\citep{shentu2024towards} employs multiple bottleneck layers of varying sizes for decoding.
% 此外已经有一些工作致力于设计Adapter在微调阶段增强基础模，如adapts,MSFT等。但是这些工作主要集中于预测任务，更重要的是他们无法帮助TSFM更好处理state variables
Additionally, some work has focused on designing adapters to augment TSFMs during fine-tuning phase, such as AdaPTS~\citep{benechehab2025adapts} and MSFT~\citep{qiao2025multi}. However, these efforts are primarily concentrated on forecasting tasks, and more importantly, fail to enable TSFMs to better process state variables.

\section{STAR}
For the task of time series anomaly detection, we consider a time series as $\boldsymbol{X} = \{ \boldsymbol{X}^n,\boldsymbol{X}^s\} \in \mathbb{R}^{T\times (C_n+C_s)}$, which consists of $C_n$ numerical variables $\boldsymbol{X}^n$ and $C_s$ state variables $\boldsymbol{X}^s$ over $T$ time points. A TSFM is leveraged to identify anomalies in a previously unseen test series $\boldsymbol{X}_{\text{test}}$ either via direct inference or subsequent fine-tuning. Specifically, the model's goal is to generate a binary prediction sequence $\boldsymbol{Y}_{\text{test}} = (y_1, y_2, \dots, y_{T_{\text{test}}})\in \mathbb{R}^{T_{\text{test}}}$. Each element $y_t \in \{0, 1\}$ in this sequence serves as a label, signifying whether the corresponding data point $y_t$ at time $t$ is anomalous.

\subsection{Overview}
\begin{figure*}[!htbp]
    \centering
\includegraphics[width=1\linewidth]{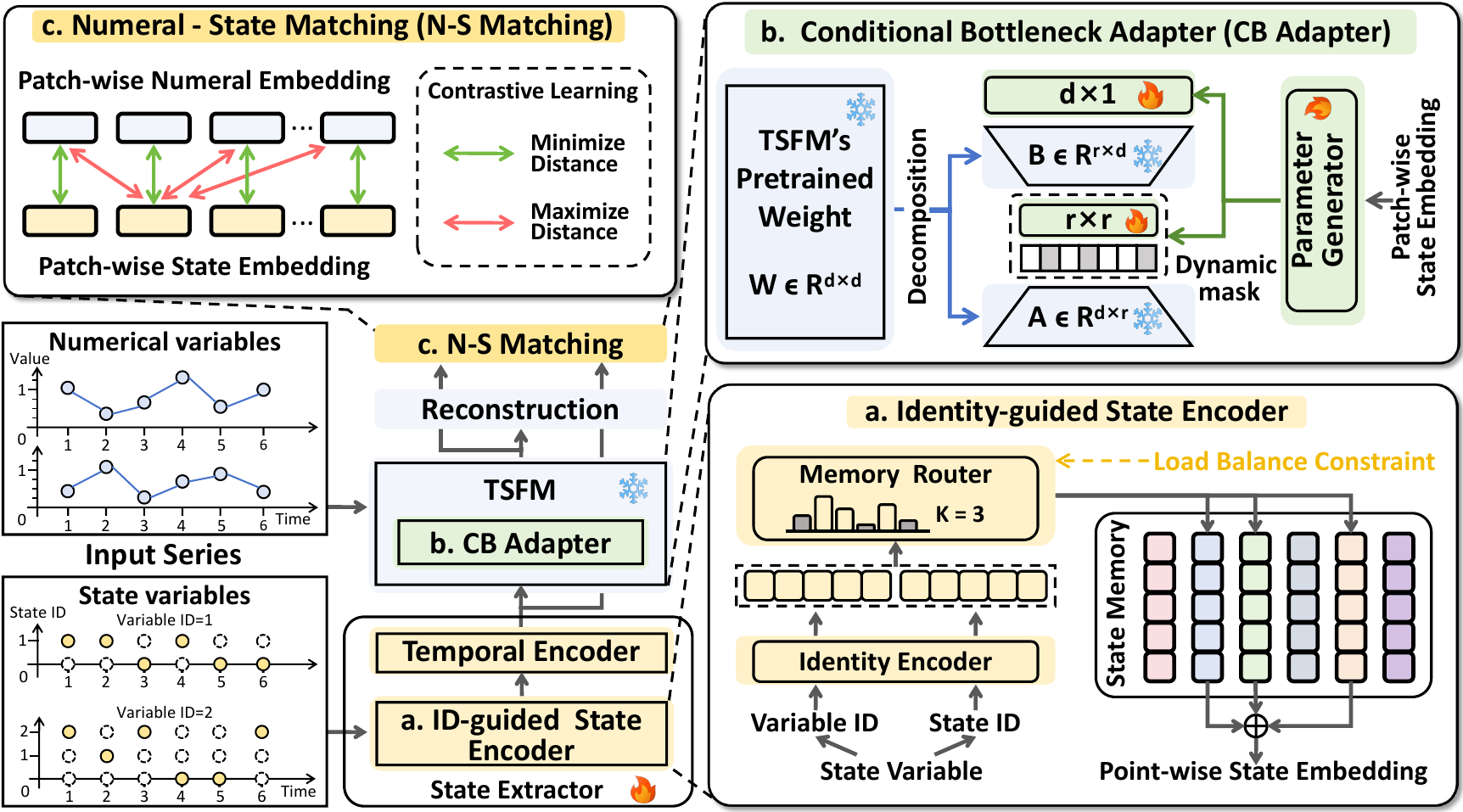}
    \caption{The overview of STAR.}
\label{fig: overview}
\vspace{-3mm}
\end{figure*}

Figure~\ref{fig: overview} shows the overall architecture of \textbf{STAR}, which reconsiders the state variables for TSFMs using the \textbf{ST}ate-aware \textbf{A}dapte\textbf{R}. We shift the processing of state variables from TSFMs to STAR. It mainly consists of three modules: 1) \textbf{State Extractor}.     Addressing the specific characteristics of state variables, this module simultaneously extracts their state and temporal features and utilizes an \textit{Identity-guided Selective State Encoder} to model the complex semantics of state variables effectively. 2) \textbf{Conditional Bottleneck Adapter}. To better model the condition-based influence, we no longer directly model interactions between state and numerical variables. Instead, we use the state variable to modulate the TSFM’s parameters, thereby influencing its processing of numerical variable.
% This module integrates state variables to construct an adapter that can dynamically adjust the bottleneck information, thereby more effectively modeling the conditional mechanism of state variables on anomaly detection. 
3) \textbf{Numeral-State Matching}. To better detect anomalies in the state, we design a contrastive learning based on the degree of matching between numerical and state variables at the patch level. Notably, during the fine-tuning process, the entire Backbone network is frozen, and only STAR is utilized to perform the fine-tuning effectively.

\subsection{State Extractor Guided by Identity}

The \textit{State Extractor} operates in two sequential stages. First, an \textit{Identity-guided State Encoder} captures state information from the variable identities (indices of state variables) and the state identities (discrete values of state variables) at each time point. Subsequently, a \textit{Temporal Encoder} extracts temporal patterns within each patch. 

\subsubsection{Identity-guided State Encoder} % 状态变量组织形式上是一组描述当前系统的具有不同类别数的类别特征。现有的工作通常对于每一个特定的类别使用可学习的向量学习其表征。这种方式会给Adapter带来庞大的参数量与学习压力，更重要的是它不能考虑到不同状态变量之间的内在联系。
Formally, state variables are a type of categorical variable, and different state variables can have different cardinality or the number of categories. Most existing methods learn a representation for every individual category via a dedicated, learnable vector to embed the categorical variable \citep{shan2016deep, hollmann2025accurate, li2025swea}. This method, however, imposes a substantial parameter burden and learning complexity. More importantly, it fails to account for the similarity or heterogeneity among different state variables. 

Inspired by the Mixture of Experts architecture \citep{zhou2022mixture, cai2025survey}, which uses router networks to select experts for different inputs, we propose a novel approach. Our approach introduces a learnable, compact State Memory, denoted as $\boldsymbol{S}\in \mathbb{R}^{N\times d}$, which consists of $N$ $d$-dimensional vectors. Under the guidance of variable identities $\boldsymbol{I}_v\in \mathbb{R}^{T \times C_s \times d}$ and state identities $\boldsymbol{I}_s\in \mathbb{R}^{T \times C_s \times d}$, a linear combination of a vector subset from this State Memory is selected to serve as the state representation. We begin by employing a Sinusoidal Encoding~\citep{DBLP:conf/nips/VaswaniSPUJGKP17} to transform the variable and state identity into $\boldsymbol{I}_v$ and $\boldsymbol{I}_s$ .
\begin{gather}
    \boldsymbol{I}_v[:,i,j]=\begin{cases} 
    \sin\left(\frac{i}{\lambda^{2k/d}}\right), & \text{if } j = 2n \\
    \cos\left(\frac{i}{\lambda^{2k/d}}\right), & \text{if } j = 2n+1 
\end{cases},~~
 \boldsymbol{I}_s[:,i,j]=\begin{cases} 
    \sin\left(\frac{s(i)}{\lambda^{2k/d}}\right), & \text{if } j = 2n \\
    \cos\left(\frac{s(i)}{\lambda^{2k/d}}\right), & \text{if } j = 2n+1 
\end{cases}
\end{gather}
Where $\lambda$ is a constant, $n$ represents any natural number, and $s(\cdot)$ defines the mapping from a state variable's identity to its state's identity at the corresponding time point. The \textit{Memory Router} leverages $\boldsymbol{I}_v$ and $\boldsymbol{I}_s$ to dynamically choose a subset of K vectors from the State Memory. To preserve a continuous gradient flow, we implement the selection mechanism using a soft mask.
\begin{gather}
    \boldsymbol{W}_{s} = f(\text{concat}(\boldsymbol{I}_v,\boldsymbol{I}_s))\in \mathbb{R}^{T \times C_s \times N},~~
    \boldsymbol{\theta} =  \text{topK}(\boldsymbol{W}_{s})\in \mathbb{R}^{T \times C_s \times 1} \\
    \boldsymbol{W}_{\text{mask}} = \text{softmax}(\boldsymbol{W}_{s} + \text{log}(\text{sigmoid}((\boldsymbol{W}_{s}-\boldsymbol{\theta})/\epsilon))) \in \mathbb{R}^{T \times C_s \times N}
\end{gather}

Where $f$, implemented as an MLP, learns the selection weights from the concatenated embeddings. The function $\text{topK}$ returns the K-th largest value, denoted as $\theta$. $\boldsymbol{W}_{\text{mask}}$ represents the weights after applying the soft mask, where all values smaller than $\boldsymbol{\theta}$ are attenuated towards zero. $\epsilon$ is a small positive constant that pushes the sigmoid's output to its extremes (0 or 1). The point-wise state embedding $\boldsymbol{S}_{\text{point}} \in \mathbb{R}^{T \times C_s \times d}$ is obtained via the equation: 
$\boldsymbol{S}_{\text{point}} = \boldsymbol{W}_{\text{mask}} \otimes \boldsymbol{S}$. 

To further promote the comprehensive training of all elements within the State Memory and to prevent routing imbalance, we introduce a constraint based on the Coefficient of Variation~\citep{lovie2005coefficient}. This constraint is applied to both the selection frequency $\boldsymbol{E}_{\text{sel}}$ and the routing importance of each element $\boldsymbol{E}_{\text{imp}}$ to ensure load balancing.
\begin{gather}
    \boldsymbol{E}_{\text{sel}} = \text{avg}(\text{sigmoid}((\boldsymbol{W}_{s}-\boldsymbol{\theta})/\epsilon))) \in \mathbb{R}^{N}, ~~\boldsymbol{E}_{\text{imp}} = \text{avg}(\boldsymbol{W}_{\text{mask}}) \in \mathbb{R}^{N}
    \\
    \mathcal{L}_{\text{bal}} = (\frac{\text{var}(\boldsymbol{E}_{\text{sel}}) }{\text{avg}(\boldsymbol{E}_{\text{sel}}) })^2 + (\frac{\text{var}(\boldsymbol{E}_{\text{imp}}) }{\text{avg}(\boldsymbol{E}_{\text{sel}}) })^2
\end{gather}
where $\text{avg}(\cdot)$, $\text{var}(\cdot)$ are functions that compute the mean and variance, respectively. $\mathcal{L}_{\text{bal}}$ denotes the load-balancing loss, which is incorporated into the overall training objective.

\subsubsection{Temporal Encoder} Motivated by the observation that state variables often exhibit temporal patterns, such as the strong periodicity of a variable representing the day of the week, we further model the temporal interactions among these embeddings across time to derive patch-wise state embeddings $\boldsymbol{S}_{\text{patch}}$. 
Initially, the point-wise state embeddings $\boldsymbol{S}_{\text{point}}$ are segmented into $m$ patches of length $l$ in a manner consistent with the backbone's configuration, yielding $\boldsymbol{P}\in \mathbb{R}^{ m \times l \times C_s \times d}$. This process comprises two components: intra-patch and inter-patch interaction.
\begin{gather}
    \boldsymbol{P}_{\text{intra}} = f_1( \boldsymbol{P})\in \mathbb{R}^{m\times C_s \times d },~~
    \boldsymbol{P}_{\text{inter}} = f_2(\boldsymbol{P}_{\text{intra}}) +  \boldsymbol{P}_{\text{intra}}\in \mathbb{R}^{m\times C_s \times d }
\end{gather}

where $f_1: \mathbb{R}^{l\times d}\to \mathbb{R}^{d}$ aggregates the representations within each patch, and $f_2: \mathbb{R}^{m}\to \mathbb{R}^{m}$ extracts valuable contextual information for each patch. Each state variable reflects only a partial aspect of the system's information. Therefore, to construct a more holistic representation of the system's overall state, we further perform an aggregation over these variables inside each patch.
\begin{gather}
    \boldsymbol{W}_{\text{agg}} = f_3(\boldsymbol{P}_{\text{inter}}) \in \mathbb{R}^{m\times C_s}, ~~\boldsymbol{S}_{\text{patch}} = f_4(\text{sum}(\boldsymbol{W}_{\text{agg}}\odot \boldsymbol{P}_{\text{inter}}))\in \mathbb{R}^{m\times d}
\end{gather}
where $f_3: \mathbb{R}^d \to \mathbb{R}^1$ is a function that adaptively computes the aggregation weights, and $f_4: \mathbb{R}^d \to \mathbb{R}^d$ serves to refine the resulting fused representation further,  $\odot$ denotes element-wise multiplication. For a lightweight implementation, both $f_1$, $f_2$, $f_3$ and $f_4$ are implemented as MLPs.

\subsection{Conditional Bottleneck Adapter}
% 得益于仅通过训练少量参数就可以完成模型微调，LoRA在广泛应用于大模型的微调阶段。为了进一步提升LoRA的效率，我们对LoRA的结构进一步优化为：
LoRA~\citep{hulora} learn a new matrix product of two low-rank matrices to modulate large-scale models' parameters. Formally, for a pretrained weight matrix $\boldsymbol{W}_0\in \mathbb{R}^{d_{\text{in}}\times d_{\text{out}}}$, the weight update $\Delta\boldsymbol{W}$ is constrained to two low-rank matrix.
This process can be formalized as:
\begin{gather}
    \boldsymbol{h} = \boldsymbol{W}_0\boldsymbol{x} + \Delta\boldsymbol{W}\boldsymbol{x}, \Delta\boldsymbol{W} = \boldsymbol{A}\boldsymbol{B}
\end{gather}
where $\boldsymbol{h}$ and $\boldsymbol{x}$ denote the input and output of the module, respectively. 
$\boldsymbol{A}\in\mathbb{R}^{d_{\text{in}}\times r}$ and $\boldsymbol{B}\in\mathbb{R}^{d_{\text{out}}\times r}$ 
are two trainable low-rank matrices, with the rank $r \ll \min(d_{\text{in}}, d_{\text{out}})$.

In a similar vein, we propose the Conditional Bottleneck Adapter. This module perceives the system's state to dynamically generate its parameters and determine its bottleneck size, as illustrated in Figure~\ref{fig: overview}b. The weight update $\Delta \boldsymbol{W}$ is given by the following equation:
    \begin{gather}
        \Delta\boldsymbol{W} = \boldsymbol{A} \boldsymbol{R}  \boldsymbol{B} \odot \boldsymbol{D},~~\boldsymbol{R}\in\mathbb{R}^{r\times r} \text{, } \boldsymbol{D}\in\mathbb{R}^{d_{\text{out}}\times 1}
    \end{gather}
    where $\boldsymbol{A}$ and $\boldsymbol{B}$ are decomposed from $\boldsymbol{W}_0$, as detailed in Section ~\ref{sec: decomposition},  and remain frozen. $\boldsymbol{R}, \boldsymbol{B}$ are determined by the state variable as detailed in Section ~\ref{sec: generator}. 

Notably, inspired by~\citep{shentu2024towards}, we employ a dynamic mask on the $\boldsymbol{R}$ matrix to modulate the information bottleneck. This mechanism enables the model to adaptively filter information based on the characteristics of diverse input series.

\subsubsection{Pretrained Weight Decomposition}
\label{sec: decomposition}
% 为了尽可能保留原始参数的性能，我们采用了奇异值分解（SVD）来计算A和B
To preserve the performance of the original parameters as much as possible, we employ Singular Value Decomposition (SVD)~\citep{stewart1993early} to decompose the pretrained weight.
\begin{gather}
    \boldsymbol{U}, \boldsymbol{\Sigma}, \boldsymbol{V} = \text{SVD}(\boldsymbol{W}_0), \boldsymbol{U}\in\mathbb{R}^{d_{\text{in}}\times d_{\text{in}}},\boldsymbol{\Sigma}\in\mathbb{R}^{d_{\text{in}}\times d_{\text{out}}} , \boldsymbol{V}\in\mathbb{R}^{d_{\text{out}}\times d_{\text{out}}}
\end{gather}
where, $\boldsymbol{\Sigma}$ is the singular value matrix, which we truncate to its leading $r$ values 
to form the matrix $\boldsymbol{\Sigma}_1\in \mathbb{R}^{d_{\text{in}}\times r}$ and $\boldsymbol{\Sigma}_2\in \mathbb{R}^{r\times d_{\text{out}}}$ for the purpose of low-rank approximation. Then we compute the low-rank  matrices as :
\begin{gather}
    \boldsymbol{A} = \boldsymbol{U}\boldsymbol{\Sigma}_1 \in \mathbb{R}^{d_{\text{in}}\times r},  \boldsymbol{B} = \boldsymbol{\Sigma}_2\boldsymbol{V} \in \mathbb{R}^{r \times d_{\text{out}}}
\end{gather}

\subsubsection{Parameter Generator and Dynamic Mask}
\label{sec: generator}
% 对于第t个patch, 我们使用其对应的S_patch来计算状态感知的参数R D.
For the $t$-th patch, we utilize its corresponding $\boldsymbol{S}_{\text{patch}}^t$ to compute the state-dependent parameters $\boldsymbol{R}$ and $\boldsymbol{D}$. It is worth noting that the $\boldsymbol{R}$ and $\boldsymbol{D}$ matrices are patch-specific, 
as they are dynamically generated based on the distinct state of each individual patch.
\begin{gather}
    \boldsymbol{R}_{\text{init}},\boldsymbol{D} = g_1(\boldsymbol{S}_{\text{patch}}^t), g_1:=\mathbb{R}^d\to \mathbb{R}^{(r\times r + d_{\text{out}})}
\end{gather}
To facilitate the adaptive adjustment of the information bottleneck, the $\boldsymbol{R}_{\text{init}}$ matrix is further subjected to a dynamic masking process.
\begin{gather}
    \boldsymbol{R}_{\text{mask}} = g_2(\boldsymbol{S}_{\text{patch}}^t)\in    \mathbb{R}^{r}, \Gamma = g_3(\boldsymbol{S}_{\text{patch}}^t)\\
    \boldsymbol{M} = \text{sigmoid}((\boldsymbol{R}_{\text{mask}} - \Gamma)/\epsilon)\in \mathbb{R}^{r\times 1}, \boldsymbol{R} = \boldsymbol{M}\odot \boldsymbol{R}_{\text{init}}
\end{gather}
% 其中\boldsymbol{R}_{\text{mask}}为r矩阵中每行生成一个分数，\Gamma代表mask阈值，M是用于软mask的矩阵，如果\boldsymbol{R}_{\text{mask}}中对应位置的元素小于\Gamma则M中的元素将趋于0.
where $g_2:=\mathbb{R}^d\to \mathbb{R}^{r}$, $g_3:=\mathbb{R}^d\to \mathbb{R}^{1}$.  $g_1,g_2,g_3$ are implemented as MLPs.
$\boldsymbol{R}_{\text{mask}}$ is score for each row in $\boldsymbol{R}$, 
$\Gamma$ is a masking threshold. $\boldsymbol{M}$ serves as the soft mask matrix, 
where if an element in $\boldsymbol{R}_{\text{mask}}$ is below the threshold $\Gamma$, 
the corresponding elements in $\boldsymbol{M}$ are driven towards zero.
\subsection{Pipeline with Numeral - State Matching}
% 异常不一定仅体现在数值变量的异常时间模式，也可能体系在状态变量的异常状态。现有的基础模型仅基于时序重构的方式检测异常，由于状态变量表现为不同的状态类别，仅重构状态类别的索引，不足以准备检测其中的异常。为了解决这一问题，我们设计了Numeral - State matching模块，通过学习数值变量与状态变量之间的匹配程度，来检测异常的发生。
Anomalies may manifest not only as aberrant temporal patterns in numerical variables but also as abnormal conditions within the state variables. Existing TSFMs, however, detect anomalies solely through time series reconstruction. Since state variables are represented as categorical variables, merely reconstructing their state identities is insufficient for accurately identifying state-based anomalies. To address this limitation, we introduce the \textit{Numeral-State Matching} module. This module is designed to detect anomalies by learning the degree of correspondence between numerical variables and their associated state variables.

\subsubsection{Finetuning phase.}
% 由于通过注入异常的方式构造正负样本对进行对比学习的方法，会为训练带来额外的开销，且难以确保注入异常的有效性。为了避免上述问题，我们选取相对应Patch-wise Numeral Embedding 和 Patch-wise State为正样本对，其余为负样本对 as shown in figure 2.
Constructing positive and negative pairs for contrastive learning by injecting anomalies introduces significant training overhead and makes it difficult to guarantee the effectiveness of the synthetic anomalies \citep{simad,kimcausality}. To circumvent these issues, we adopt a different approach. As shown in Figure~\ref{fig: overview}c, we define a positive pair as the corresponding patch-wise numeral embedding $\boldsymbol{N}_{\text{patch}}^t \in \mathbb{R}^{d}$ and patch-wise state embedding $\boldsymbol{S}_{\text{patch}}^t \in \mathbb{R}^{d}$. All other non-corresponding pairings are consequently treated as negative examples. The loss for the Numeral - State Matching Contrastive Learning can be expressed as follows:
\begin{gather}
       \mathcal{L}_{\text{match}} = -\frac{1}{M}\sum_{t=1}^M\text{log}(\frac{ \text{exp}(\text{sim}(\boldsymbol{N}_{\text{patch}}^t,\boldsymbol{S}_{\text{patch}}^t)/\tau)}{\sum_{k=1}^M  \text{exp}(\text{sim}(\boldsymbol{N}_{\text{patch}}^k,\boldsymbol{S}_{\text{patch}}^t)/\tau))})
\end{gather}
% where $m$ 是patch的总数，$\boldsymbol{N}_{\text{patch}}^i, \boldsymbol{S}_{\text{patch}}^i$ 分别代表第i个patch对应的numeral embedding和state embedding，$\text{sim}$指代余弦相似度， $\tau$是控制对比学习程度的温度系数。
where, $M$ denotes the total number of patches in one batch.  $\boldsymbol{N}_{\text{patch}}^i$ and $\boldsymbol{S}_{\text{patch}}^i$ are the respective numeral and state embeddings corresponding to the $i$-th patch. The function $\text{sim}(\cdot)$ is the cosine similarity, and $\tau$ is a temperature coefficient used to scale the logits in the contrastive loss.

The overall loss during training is composed of three components: the numerical variable reconstruction loss ($\mathcal{L}_{\text{rec}}$), the load-balancing loss ($\mathcal{L}_{\text{balance}}$), and the Numeral-State Matching loss ($\mathcal{L}_{\text{match}}$). The total loss, $\mathcal{L}_{\text{total}}$, is formulated as:
$\mathcal{L}_{\text{total}} = \mathcal{L}_{\text{rec}} + \lambda_1 \mathcal{L}_{\text{balance}} + \lambda_2 \mathcal{L}_{\text{match}}$.
where $\lambda_1$ and $\lambda_2$ are hyperparameters that balance the contribution of each loss term.

\subsubsection{Inference phase}
% 异常检测由两部分组成：点级别的数值重构以及patch级别的NUMERAL - STATE MATCHING. 在测试时间序列序列中分别计算他们的得分$score_rec$, $score_match$
Anomaly detection is performed in two parts: point-wise \textit{numerical reconstruction} and patch-wise \textit{Numeral-State Matchin}g. During inference on a test time series, their respective scores, denoted as $\text{score}_\text{rec}$
and $\text{score}_{\text{match}}$, are calculated separately.
\begin{gather}
    \boldsymbol{\text{score}}_{rec} = \text{avg}(\text{mse}(\hat{\boldsymbol{X}}_{\text{test}}^n,\boldsymbol{X}_{\text{test}}^n)) \in \mathbb{R}^{T_{\text{test}}},~~\boldsymbol{\text{score}}_{rec} = \text{sim}(\boldsymbol{N}_{\text{patch}},\boldsymbol{S}_{\text{patch}}) \in \mathbb{R}^{M_{\text{test}}}
\end{gather}
where $\text{mse}(\cdot)$ is the Mean Square Error, $\text{sim}(\cdot)$ denotes the cosine similarity, $\hat{\boldsymbol{X}}^n$ represents the reconstruction output of numerical variables from the backbone, and $T_{\text{test}}$ and $m_{\text{test}}$ are the total number of time points and patches in the test set, respectively. avg(·) is an operation that computes the mean value across the variable dimension.

Given that $\text{score}_\text{rec}$ and $\text{score}_{\text{match}}$ differ in both temporal granularity and numerical scale, we first apply linear interpolation to  $\text{score}_{\text{match}}$ to align their resolutions. Subsequently, the two scores are fused by taking their element-wise product.
\begin{gather}
    \boldsymbol{\text{score}}_{\text{match}}' = \text{Linear-Interpolation}(\boldsymbol{\text{score}}_{\text{match}}) \in \mathbb{R}^{T_{\text{test}}}\\
    \boldsymbol{\text{score}}_{\text{total}} = \boldsymbol{\text{score}}_{rec} \times \text{softmax}(\boldsymbol{\text{score}}_{\text{match}}') \in \mathbb{R}^{T_{\text{test}}}
\end{gather}

% 我们的方法不仅可以检测数值变量的变化异常，更可以感知系统状态的波动异常，同时，得益于patch级别的对比学习，也使得模型可以更好地检测子序列异常。
Our approach not only detects anomalies in numerical variables but also perceives anomalous in the system state. Furthermore, by leveraging patch-wise similarity computation, the model's capability to detect subsequence anomalies is effectively enhanced.

\section{Experiments}
\subsection{Experimental Settings}
\label{sec: exp setting}
% \textbf{Datasets} We evaluate STAR on various datasets with \textit{state  variables}. Here is the description of these datasets: 1) \textbf{MSL} (Mars Science Laboratory Dataset), collected by NASA, includes telemetry data that reflects the operational status of sensors and actuators on the Martian rover~\citep{msl}. 2) \textbf{SMAP} (Soil Moisture Active Passive Dataset), also gathered by NASA, provides soil moisture data obtained from spacecraft monitoring systems~\citep{msl}. 3) \textbf{SWaT} (Secure Water Treatment) contains sensor data from a continuously operating water treatment infrastructure~\citep{swat}. 4) \textbf{Genesis} A sensor and control signals dataset collected from cyber-physical production systems~\citep{Genesis}. 5) \textbf{NYC} The Transportation dataset provides information on taxi and ride-hailing trips in New York. The statistical details about the datasets are available in Appendix~\ref{app: datasets}.

\textbf{Datasets} We evaluate STAR on various datasets with state  variables, including \textbf{MSL} (Mars Science Laboratory Dataset)~\citep{msl},  \textbf{SMAP} (Soil Moisture Active Passive Dataset)~\citep{msl}, \textbf{SWaT} (Secure Water Treatment)~\citep{swat}, \textbf{Genesis}~\citep{Genesis}, and \textbf{NYC}~\citep{nyc}. The statistical details about the datasets are available in Appendix~\ref{app: datasets}.

\textbf{Backbone} We choose the latest state-of-the-art TSFMs for anomaly detection as backbones, including DADA~\citep{shentu2024towards}, UniTS~\citep{units}, Moment~\citep{goswami2024moment}, and Timer~\citep{timer}. The detailed description is available in Appendix~\ref{app: backbones}. 

\textbf{Metrics} To circumvent the potential for unfair comparisons arising from the diverse threshold selection methods employed by different TSFMs, we primarily adopt threshold-irrelevant metrics for evaluation. These include the AUC-ROC (A-R), VUS-ROC (V-R), and VUS-PR (V-P)~\citep{paparrizos2022volume}. Meanwhile, the comprehensive evaluation is also supplemented by a range of commonly used metrics mentioned in TAB\citep{qiu2025tab}. The description and implementation details of all metrics are shown in Appendix~\ref{app: Metrics}.
\subsection{Detection results}
\textbf{Main results} We evaluate our model on five real-world datasets containing state  variables, using four TSFMs as backbones. We uniformly use 10\% of the downstream data for fine-tuning. The results are summarized in Table~\ref{tab: main result}. For a fair comparison, we evaluated two baseline fine-tuning methods: (1) \textbf{Backbone}: standard fine-tuning, and (2) \textbf{LoRA}: fine-tuning the base model with LoRA~\citep{hulora}. It can be seen that LoRA-based fine-tuning offers inconsistent and often marginal performance changes, our method consistently delivers substantial enhancements to the detection performance of TSFMs. Compared to standard fine-tuning, STAR improves the V-R metric on task-specific and task-general TSFMs by 6.93\% and 6.07\%, respectively. This demonstrates the generalizability of STAR across different types of TSFMs. 
The validity of our motivation and modeling is further highlighted on the Genesis and NYC, where STAR delivers more improvements of 8.73\% and 9.17\%, owing to the meaningful physical interpretations of their state  variables. 
% Notably, despite the significant performance discrepancies on the SWaT dataset, this variance is considered to be within a reasonable range due to the unique characteristics of its distribution. A similar phenomenon can be found in~\cite{wucatch}.

\textbf{Multi-metrics} To ensure a comprehensive comparison, we evaluate the models on more metrics, including Accuracy (ACC), Range-Precision (R-P), Range-Recall (R-R), Range-F1-score (R-F1)~\citep{tatbul2018precision}, Affiliated-Precision (Aff-P), Affiliated-Recall (Aff-R), Affiliated-F1 (Aff-f1)~\citep{huet2022local}, AUC-PR (A-P), R-AUC-PR (R-A-P), R-AUC-ROC (R-A-R)~\citep{paparrizos2022volume}. Table~\ref{tab: full metrics} shows that STAR yields performance improvements across the majority of metrics.
\begin{table}[!t]

    \caption{ Results of adapters for different TFSMs. The better results are highlighted in \textbf{bold}. }
    \centering
    \resizebox{1\linewidth}{!}{
    \setlength{\tabcolsep}{3pt}
    \begin{tabular}{l|c c c|c c c|c c c|c c c|c c c}
    \specialrule{0.8pt}{0pt}{2pt}
        \textbf{Dataset} & \multicolumn{3}{c|}{\textbf{MSL}} & \multicolumn{3}{c|}{\textbf{SMAP}}  & \multicolumn{3}{c|}{\textbf{SWaT}} & \multicolumn{3}{c|}{\textbf{Genesis}}& \multicolumn{3}{c}{\textbf{NYC}} \\ \midrule
         \textbf{Metric} & \textbf{A-R} & \textbf{V-R} & \textbf{V-P} & \textbf{A-R} & \textbf{V-R} & \textbf{V-P} & \textbf{A-R} & \textbf{V-R} & \textbf{V-P} & \textbf{A-R} & \textbf{V-R} & \textbf{V-P} & \textbf{A-R} & \textbf{V-R} & \textbf{V-P} \\ \specialrule{0.8pt}{2pt}{2pt}
        \textbf{DADA} &0.702          & 0.746          & 0.284          & 0.406          & 0.451          & 0.123          & 0.814          & 0.740          & 0.558          & 0.802          & 0.832          & 0.154          & 0.492          & 0.652          & 0.050          \\\midrule
 
        \textcolor{darkgreen}{\textbf{~+LoRA}} &0.721          & 0.765          & 0.298          & 0.420          & 0.463          & 0.129          & 0.807          & 0.735          & 0.562          & 0.784          & 0.818          & 0.153          & 0.514          & 0.666          & 0.052          \\\midrule
        
        \textcolor{redorange}{\textbf{~+STAR}} &\textbf{0.787} & \textbf{0.803} & \textbf{0.323} & \textbf{0.447} & \textbf{0.490} & \textbf{0.129} & \textbf{0.838} & \textbf{0.753} & \textbf{0.577} & \textbf{0.896} & \textbf{0.911} & \textbf{0.165} & \textbf{0.571} & \textbf{0.701} & \textbf{0.065} \\ \specialrule{0.8pt}{2pt}{2pt}
        
        \textbf{UniTS} & 0.747          & 0.794          & 0.296          & 0.515          & 0.555          & 0.145          & 0.236          & 0.337          & 0.126          & 0.704          & 0.770          & 0.017          & 0.486          & 0.613          & 0.044          \\\midrule
         
        \textcolor{darkgreen}{\textbf{~+LoRA}} &0.748          & 0.796          & 0.298          & 0.529          & 0.567          & 0.148          & 0.235          & 0.336          & 0.125 & 0.706          & 0.777          & 0.018          & 0.508          & 0.622          & 0.049          \\\midrule
        
        \textcolor{redorange}{\textbf{~+STAR}} &\textbf{0.776} & \textbf{0.810} & \textbf{0.332} & \textbf{0.535} & \textbf{0.575} & \textbf{0.154} & \textbf{0.242} & \textbf{0.355} & \textbf{0.137} & \textbf{0.809} & \textbf{0.850} & \textbf{0.038} & \textbf{0.595} & \textbf{0.717} & \textbf{0.082} \\ \specialrule{0.8pt}{2pt}{2pt}
        
        \textbf{Moment} & 0.714          & 0.753          & 0.281          & 0.520          & 0.561          & 0.145          & 0.267          & 0.376          & 0.185          & 0.886          & 0.906          & 0.155          & 0.521          & 0.632          & 0.045          \\\midrule
         
        \textcolor{darkgreen}{\textbf{~+LoRA}} &0.716          & 0.755          & 0.287          & 0.521          & 0.563          & 0.145          & 0.272          & 0.383          & 0.190          & 0.889          & 0.908          & 0.184          & 0.526          & 0.636          & 0.045          \\\midrule
        
        \textcolor{redorange}{\textbf{~+STAR}} &\textbf{0.771} & \textbf{0.807} & \textbf{0.309} & \textbf{0.524} & \textbf{0.562} & \textbf{0.146} & \textbf{0.283} & \textbf{0.388} & \textbf{0.198} & \textbf{0.949} & \textbf{0.950} & \textbf{0.254} & \textbf{0.576} & \textbf{0.680} & \textbf{0.050} \\ \specialrule{0.8pt}{2pt}{2pt}
        
        \textbf{Timer} & 0.749          & 0.793          & 0.299          & 0.530          & 0.571          & 0.149          & 0.226          & 0.326          & 0.125          & 0.842          & 0.854          & 0.100          & 0.517          & 0.628          & 0.043          \\\midrule
         
        \textcolor{darkgreen}{\textbf{~+LoRA}} &0.750          & 0.789          & 0.311          & 0.535          & 0.575          & 0.155          & 0.234          & 0.337          & 0.126          & 0.902          & 0.887          & 0.158          & 0.455          & 0.602          & 0.039          \\\midrule
        
        \textcolor{redorange}{\textbf{~+STAR}} &\textbf{0.786} & \textbf{0.812} & \textbf{0.344} & \textbf{0.538} & \textbf{0.579} & \textbf{0.154} & \textbf{0.244} & \textbf{0.352} & \textbf{0.129} & \textbf{0.931} & \textbf{0.944} & \textbf{0.179} & \textbf{0.541} & \textbf{0.658} & \textbf{0.050} \\ \specialrule{0.8pt}{2pt}{0  pt} 
        
    \end{tabular}
    \label{tab: main result}
   
    }
    \vspace{-5mm}
\end{table}
\begin{table}[!ht]

    \caption{ Multi-metrics results on real-world datasets. The better results are highlighted in \textbf{bold}.}
    \centering
    \resizebox{1\linewidth}{!}{
    \setlength{\tabcolsep}{3.5pt}
    \begin{tabular}{c | c | c c c c c c c | c c c c c c}
    \specialrule{0.8pt}{0pt}{2pt}
    \multirow{2.5}{*}{\textbf{Dataset}} & \multirow{2.5}{*}{\textbf{Method}} & \multicolumn{7}{c|}{\textbf{Threshold-relevant}} & \multicolumn{6}{c}{\textbf{Threshold-irrelevant}} \\ \cmidrule{3-15}
    &   & \textbf{ACC} & \textbf{ R-P} & \textbf{R-R} & \textbf{R-F1} & \textbf{Aff-P}& \textbf{Aff-R} & \textbf{Aff-F1} & \textbf{A-R} & \textbf{A-P} & \textbf{R-A-R} & \textbf{R-A-P} & \textbf{V-R} & \textbf{V-P} \\ \specialrule{0.8pt}{2pt}{2pt}

    \multirow{5.5}{*}{\textbf{MSL}} 
    & \textbf{DADA}  & 0.814          & \textbf{0.179} & 0.288          & 0.221          & 0.603          & \textbf{0.981} & 0.747          & 0.702          & 0.232          & 0.752          & 0.290          & 0.746          & 0.284        \\\cmidrule{2-15}
    & \textcolor{redorange}{\textbf{~+STAR}}  & \textbf{0.822} & 0.164          & \textbf{0.346} & \textbf{0.223} & \textbf{0.694} & 0.952          & \textbf{0.803} & \textbf{0.790} & \textbf{0.276} & \textbf{0.811} & \textbf{0.337} & \textbf{0.807} & \textbf{0.331}   \\ \cmidrule{2-15}
    & \textbf{UniTS}  & 0.747          & \textbf{0.197} & 0.261          & 0.224          & 0.649          & \textbf{0.996} & 0.786          & 0.747          & 0.222          & 0.802          & 0.303          & 0.794          & 0.296    \\ \cmidrule{2-15}
    & \textcolor{redorange}{\textbf{~+STAR}}  & \textbf{0.816} & 0.185          & \textbf{0.337} & \textbf{0.239} & \textbf{0.692} & 0.955          & \textbf{0.803} & \textbf{0.776} & \textbf{0.255} & \textbf{0.815} & \textbf{0.341} & \textbf{0.810} & \textbf{0.332}   \\\specialrule{0.8pt}{2pt}{2pt}
    
    \multirow{5.5}{*}{\textbf{Genesis}} 
    & \textbf{DADA}  & 0.987          & 0.333          & 0.162          & 0.218          & 0.835          & 0.953          & 0.890          & 0.802          & \textbf{0.175} & 0.844          & 0.152          & 0.832          & 0.154  \\\cmidrule{2-15}
    & \textcolor{redorange}{\textbf{~+STAR}}  & \textbf{0.996} & \textbf{0.571} & \textbf{0.172} & \textbf{0.264} & \textbf{0.895} & \textbf{0.969} & \textbf{0.930} & \textbf{0.896} & 0.159          & \textbf{0.921} & \textbf{0.156} & \textbf{0.911} & \textbf{0.165}   \\ \cmidrule{2-15}
    & \textbf{UniTS}  & 0.977          & 0.031          & \textbf{0.165} & 0.058          & 0.704          & \textbf{0.976} & 0.818          & 0.704          & 0.017          & 0.791          & 0.018          & 0.770          & 0.017   \\ \cmidrule{2-15}
    & \textcolor{redorange}{\textbf{~+STAR}}  & \textbf{0.995} & \textbf{0.154} & 0.156          & \textbf{0.155} & \textbf{0.807} & 0.949          & \textbf{0.872} & \textbf{0.809} & \textbf{0.047} & \textbf{0.861} & \textbf{0.039} & \textbf{0.850} & \textbf{0.038}   \\\specialrule{0.8pt}{2pt}{2pt}
    
    \end{tabular}
    \label{tab: full metrics}
    }
\end{table}
\subsection{Model Analysis}

\textbf{Ablation Study} To analyze the contribution of each module, we conducted an ablation study, detailed in Table~\ref{tab: ablation studies}. Row 1 shows the standard fine-tuning, while Rows 2-6 illustrate the performance when integrating specific modules of STAR with the backbone. We have the following observations: (1) A comparison of rows 1-3 reveals that both the \textit{Conditional Bottleneck Adapter} (\textit{CB Adapter}) and the \textit{Numeral-State Matching} (\textit{N-S Matching}) module can enhance the backbone's performance to a certain degree. This limited improvement can likely be attributed to the lack of meaningful state information. (2) In Rows 4 and 5, the addition of the \textit{ID-guided State Encoder} results in a more pronounced performance enhancement with the two modules. This validates the effectiveness of the \textit{ID-guided State Encoder} in capturing complex state information. (3) The combination of all three modules creates a virtuous cycle, resulting in the best performance, as shown in Row 6. Specifically, the \textit{N-S Matching} module optimizes the state representations, which in turn enhances the effectiveness of the other modules.
\begin{table}[!t]

    \caption{ Ablation studies for STAR. The better results are highlighted in \textbf{bold}.}
    \centering
    \resizebox{1\linewidth}{!}{
    \label{tab: ablation studies}
    \setlength{\tabcolsep}{6pt}
    \begin{tabular}{ c c c c | c c | c c | c c | c c}
    \specialrule{0.8pt}{0pt}{2pt}
    \multicolumn{4}{c|}{\textbf{Dataset}} & \multicolumn{4}{c|}{\textbf{MSL}} & \multicolumn{4}{c}{\textbf{Genesis}}\\ \cmidrule{1-12}
     & \multirow{2.5}{*}{\makecell[c]{\textbf{ID-guided}\\\textbf{State Encoder}}} &  \multirow{2.5}{*}{\textbf{CB Adapter}} & \multirow{2.5}{*}{\makecell[c]{\textbf{N - S}\\\textbf{Matching}}} & \multicolumn{2}{c|}{\textbf{DADA}} & \multicolumn{2}{c|}{\textbf{UniTS}} &  \multicolumn{2}{c|}{\textbf{DADA}} & \multicolumn{2}{c}{\textbf{UniTS}}\\ \cmidrule{5-12}
 
     & & & &\textbf{V-R} & \textbf{V-P} & \textbf{V-R} & \textbf{V-P}& \textbf{V-R} & \textbf{V-P}& \textbf{V-R} & \textbf{V-P}\\  \specialrule{0.8pt}{2pt}{2pt}

    1 &  \without  & \without & \without & 0.746 & 0.284 & 0.794 & 0.296 & 0.832 & 0.154 & 0.770 & 0.017  \\   \midrule
    2 &  \without  & \with & \without & 0.762 & 0.295 & 0.792 & 0.299 & 0.855 & 0.154 & 0.784 & 0.022 \\   \midrule
    3 &  \without  & \without & \with & 0.755 & 0.290 & 0.800 & 0.298 & 0.841 & 0.157 & 0.780 & 0.019 \\   \midrule
    4 &  \with  & \with & \without & 0.803 & 0.323 & 0.807 & 0.311 & 0.872 & 0.156 & 0.819 & 0.021\\   \midrule
    5 &  \with  & \without & \with &0.760 & 0.295 & 0.805 & 0.321 & 0.854 & 0.150 & 0.805 & 0.024\\  \midrule
    6 &  \with  & \with & \with & \textbf{0.807} & \textbf{0.331} & \textbf{0.810} & \textbf{0.332} & \textbf{0.911} & \textbf{0.165} & \textbf{0.850} & \textbf{0.038}\\  \specialrule{0.8pt}{2pt}{2pt}
    \end{tabular}
    }
     \begin{tablenotes}
        \fontsize{9}{2}\selectfont
        \item ~~\without~~indicates a module removed, and \with~~indicates a module added.
    \end{tablenotes}

\end{table}

\begin{figure*}[t]
  \centering
  \subfloat[Number of selection]
  {\includegraphics[width=0.246\textwidth]{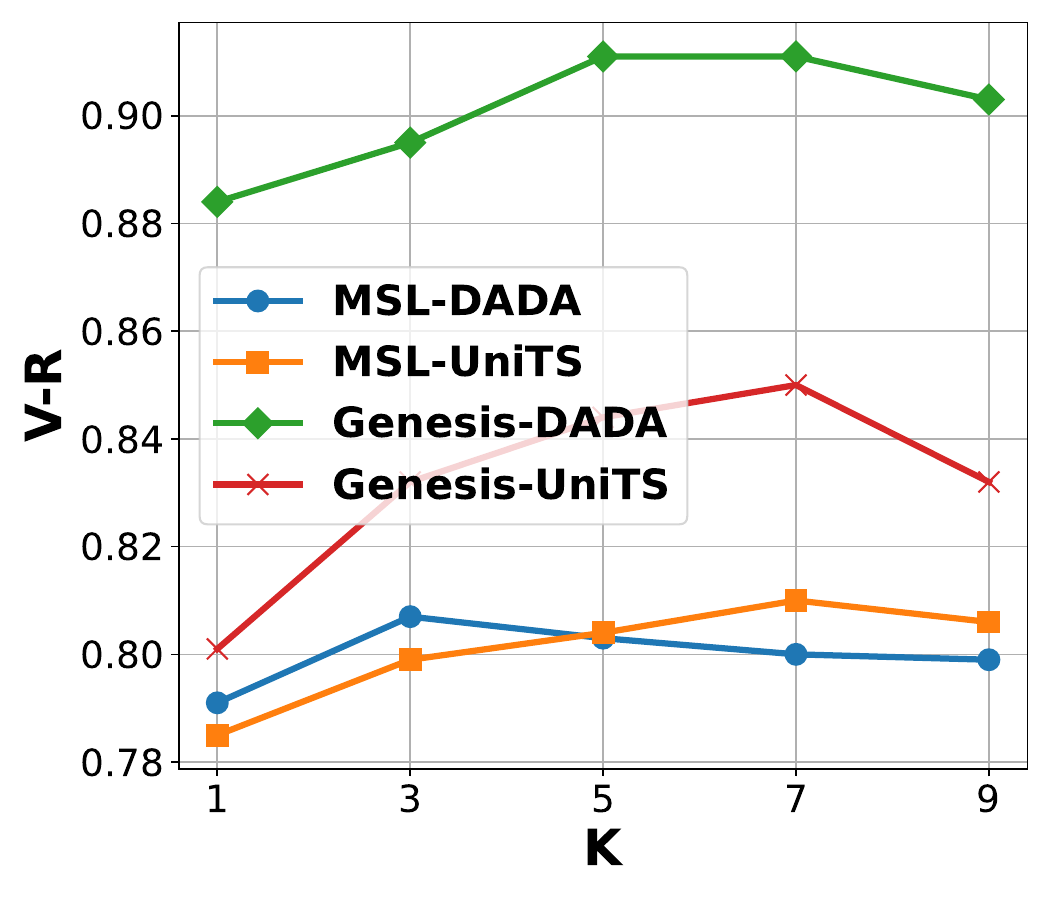}\label{fig: sen1}}  
  \subfloat[Number of memory]
  {\includegraphics[width=0.246\textwidth]{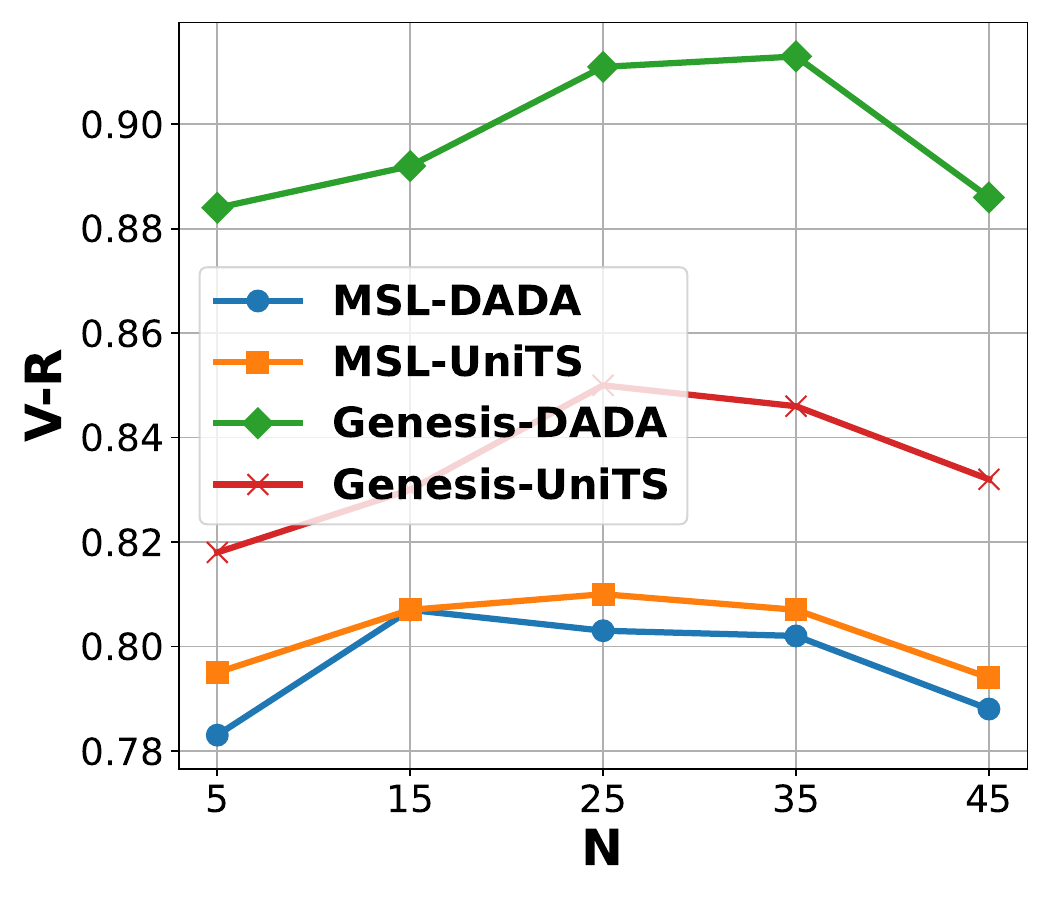}\label{fig: sen2}}  
  \subfloat[Rank of adapter]
  {\includegraphics[width=0.246\textwidth]{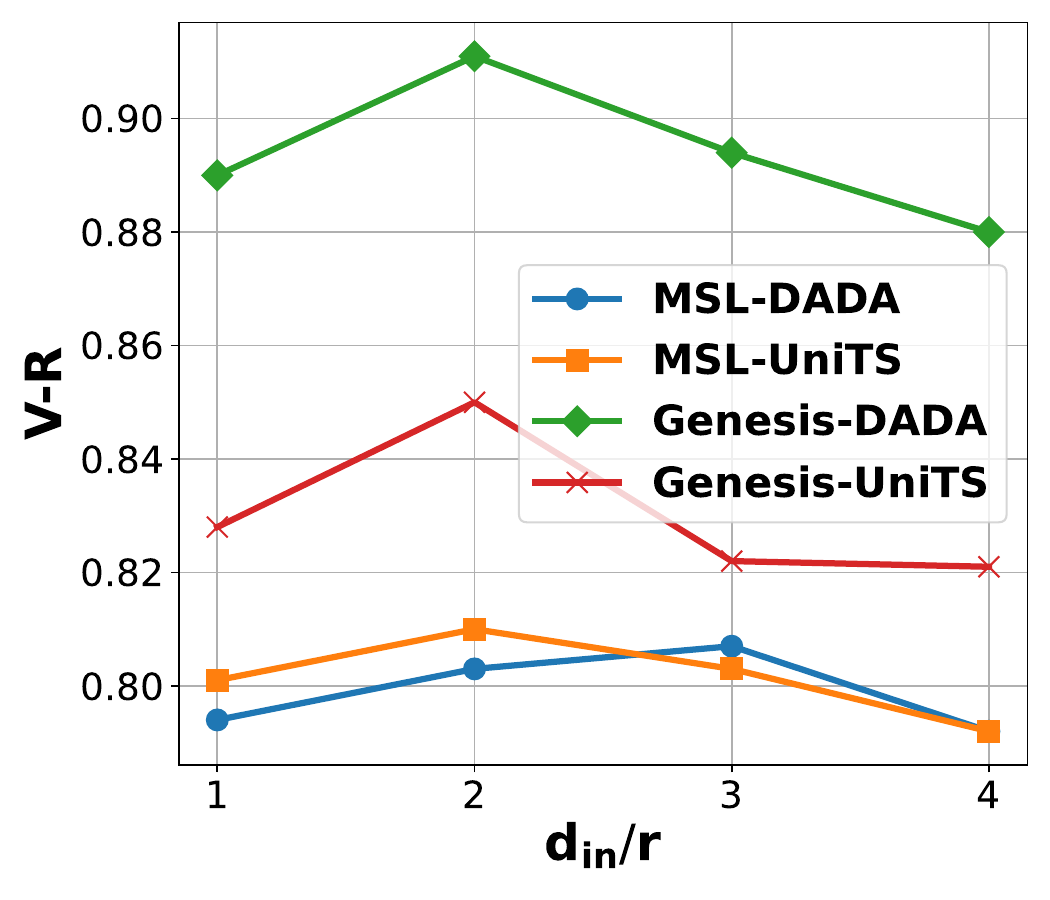}\label{fig: sen3}}
 \subfloat[Hidden dimension size]
  {\includegraphics[width=0.246\textwidth]{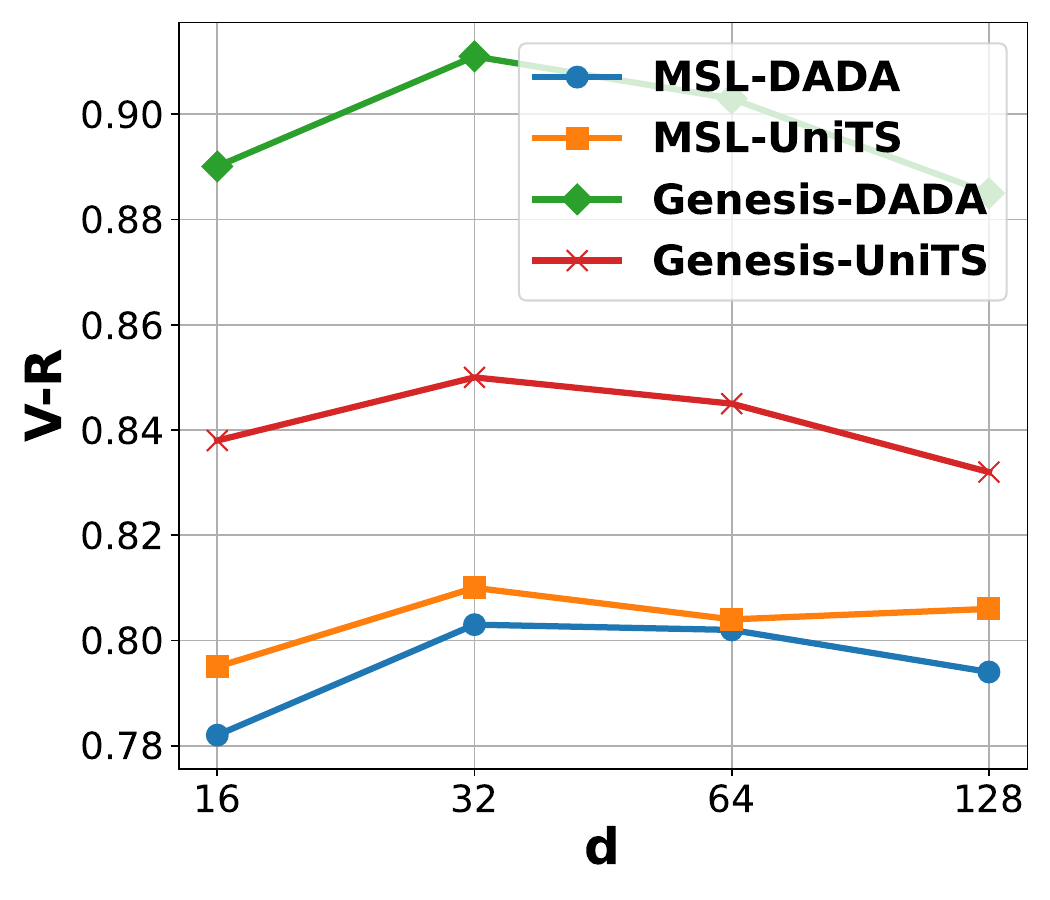}\label{fig: sen4}}
  \caption{Parameter sensitivity studies of main hyper-parameters in STAR.}
  \vspace{-4mm}
\end{figure*}

\textbf{Parameter Sensitivity}
We study the parameter sensitivity of STAR, including the number of selected vectors in \textit{Memory Router} ($K$), the total number of memory vectors ($N$), the rank of the adapter ($r$), and the hidden dimension size of the adapter ($d$). To enhance the flexibility of STAR, we search for the value of $d_{in}/r$ directly, rather than searching for $r$ itself. Figure~\ref{fig: sen1} shows the impact of $K$. In most cases, a smaller $K$ is insufficient for effective learning. We typically select 7 as the preferred option. Besides, Figure~\ref{fig: sen2} illustrates the impact of $N$. Common choices for this value are 15 or 25. This approach is more lightweight compared to defining a learnable vector for each state of each state  variable~\citep{shan2016deep}. As shown in Figure~\ref{fig: sen3}, by reducing the $d_{in}$ to $d_{in}/2$ or $d_{in}/3$, we can achieve a balance between efficiency and effectiveness. Figure~\ref{fig: sen4} indicates that a large hidden dimension is not required, thus avoiding substantial additional computational costs.

\textbf{More Analysis} To demonstrate the applicability of STAR, we also evaluated its performance on datasets containing solely numerical variables (\ref{app: Applicability for dataset}). We further visualize the \textit{Identity-guided State Encoder} to validate its effectiveness (\ref{app: Visualization}). Moreover, we present a case study to demonstrate the importance of state  variables for anomaly detection (\ref{app: case study}).

\section{Conclusion}
In this study, we propose a novel STate-aware AdapteR (STAR) for TSFM in MTSAD. STAR is a plug-and-play module designed to address the failure of TSFMs to handle state variables during the fine-tuning stage. We propose Identity-guided State Encoder and Conditional Bottleneck Adapter to capture the complex semantics of states and the conditional influence of state variables, respectively. 
% Our implementation is available at \href{https://anonymous.4open.science/r/STAR-4540}{https://anonymous.4open.science/r/STAR-4540}.

\clearpage

% \section*{Ethics statement}
% Our work exclusively uses publicly available benchmark datasets that contain no personally identifiable information. The proposed adapter for Time Series Foundation Models in Multivariate Time Series Anomaly Detection is designed for beneficial applications in system reliability and safety monitoring. No human subjects were involved in this research.

% \section*{Reproducibility statement}
% The performance of STAR and the datasets used in our work are real, and all experimental results can be reproduced. We have released our model code in an anonymous repository:
% \href{https://anonymous.4open.science/r/STAR-4540}{https://anonymous.4open.science/r/STAR-4540}. Once the paper is accepted, we will release the scripts for all settings.
\bibliography{reference}

\begin{thebibliography}{74}
\providecommand{\natexlab}[1]{#1}
\providecommand{\url}[1]{\texttt{#1}}
\expandafter\ifx\csname urlstyle\endcsname\relax
  \providecommand{\doi}[1]{doi: #1}\else
  \providecommand{\doi}{doi: \begingroup \urlstyle{rm}\Url}\fi

\bibitem[Abdulaal et~al.(2021)Abdulaal, Liu, and Lancewicki]{abdulaal2021practical}
Ahmed Abdulaal, Zhuanghua Liu, and Tomer Lancewicki.
\newblock Practical approach to asynchronous multivariate time series anomaly detection and localization.
\newblock In \emph{Proceedings of the 27th ACM SIGKDD conference on knowledge discovery \& data mining}, pp.\  2485--2494, 2021.

\bibitem[Benechehab et~al.(2025)Benechehab, Feofanov, Paolo, Thomas, Filippone, and K{\'e}gl]{benechehab2025adapts}
Abdelhakim Benechehab, Vasilii Feofanov, Giuseppe Paolo, Albert Thomas, Maurizio Filippone, and Bal{\'a}zs K{\'e}gl.
\newblock Adapts: Adapting univariate foundation models to probabilistic multivariate time series forecasting.
\newblock \emph{arXiv preprint arXiv:2502.10235}, 2025.

\bibitem[Breunig et~al.(2000)Breunig, Kriegel, Ng, and Sander]{breunig2000lof}
Markus~M Breunig, Hans-Peter Kriegel, Raymond~T Ng, and J{\"o}rg Sander.
\newblock {LOF}: identifying density-based local outliers.
\newblock In \emph{Proceedings of the 2000 ACM SIGMOD international conference on Management of data}, pp.\  93--104, 2000.

\bibitem[Cai et~al.(2025)Cai, Jiang, Wang, Tang, Kim, and Huang]{cai2025survey}
Weilin Cai, Juyong Jiang, Fan Wang, Jing Tang, Sunghun Kim, and Jiayi Huang.
\newblock A survey on mixture of experts in large language models.
\newblock \emph{IEEE Transactions on Knowledge and Data Engineering}, 2025.

\bibitem[Chen et~al.(2025{\natexlab{a}})Chen, Zhao, Song, and Xie]{chen2025unified}
Junhao Chen, Chunhui Zhao, Pengyu Song, and Min Xie.
\newblock Unified low-dimensional subspace analysis of continuous and binary variables for industrial process monitoring.
\newblock \emph{IEEE Transactions on Cybernetics}, 2025{\natexlab{a}}.

\bibitem[Chen et~al.(2025{\natexlab{b}})Chen, Huang, Cheng, Chen, Rao, Shu, Yang, Pan, and Guo]{AimTS}
Yuxuan Chen, Shanshan Huang, Yunyao Cheng, Peng Chen, Zhongwen Rao, Yang Shu, Bin Yang, Lujia Pan, and Chenjuan Guo.
\newblock {AimTS}: Augmented series and image contrastive learning for time series classification.
\newblock In \emph{ICDE}, 2025{\natexlab{b}}.

\bibitem[Cui et~al.(2016)Cui, Surpur, Ahmad, and Hawkins]{nyc}
Yuwei Cui, Chetan Surpur, Subutai Ahmad, and Jeff Hawkins.
\newblock A comparative study of htm and other neural network models for online sequence learning with streaming data.
\newblock In \emph{IJCNN}, pp.\  1530--1538, 2016.

\bibitem[Das et~al.(2024)Das, Kong, Sen, and Zhou]{timesfm}
Abhimanyu Das, Weihao Kong, Rajat Sen, and Yichen Zhou.
\newblock A decoder-only foundation model for time-series forecasting.
\newblock In \emph{Forty-first International Conference on Machine Learning}, 2024.

\bibitem[Davis \& Goadrich(2006)Davis and Goadrich]{davis2006relationship}
Jesse Davis and Mark Goadrich.
\newblock The relationship between precision-recall and roc curves.
\newblock In \emph{Proceedings of the international conference on machine learning}, pp.\  233--240, 2006.

\bibitem[Deng \& Hooi(2021)Deng and Hooi]{deng2021graph}
Ailin Deng and Bryan Hooi.
\newblock Graph neural network-based anomaly detection in multivariate time series.
\newblock In \emph{Proceedings of the AAAI conference on artificial intelligence}, volume~35, pp.\  4027--4035, 2021.

\bibitem[Ekambaram et~al.(2024)Ekambaram, Jati, Dayama, Mukherjee, Nguyen, Gifford, Reddy, and Kalagnanam]{tinyttm}
Vijay Ekambaram, Arindam Jati, Pankaj Dayama, Sumanta Mukherjee, Nam Nguyen, Wesley~M Gifford, Chandra Reddy, and Jayant Kalagnanam.
\newblock Tiny time mixers (ttms): Fast pre-trained models for enhanced zero/few-shot forecasting of multivariate time series.
\newblock \emph{Advances in Neural Information Processing Systems}, 37:\penalty0 74147--74181, 2024.

\bibitem[Fawcett(2006)]{fawcett2006introduction}
Tom Fawcett.
\newblock An introduction to roc analysis.
\newblock \emph{Pattern recognition letters}, 27\penalty0 (8):\penalty0 861--874, 2006.

\bibitem[Feng \& Tian(2021)Feng and Tian]{feng2021time}
Cheng Feng and Pengwei Tian.
\newblock Time series anomaly detection for cyber-physical systems via neural system identification and bayesian filtering.
\newblock In \emph{Proceedings of the 27th ACM SIGKDD conference on knowledge discovery \& data mining}, pp.\  2858--2867, 2021.

\bibitem[Gao et~al.(2024)Gao, Koker, Queen, Hartvigsen, Tsiligkaridis, and Zitnik]{units}
Shanghua Gao, Teddy Koker, Owen Queen, Thomas Hartvigsen, Theodoros Tsiligkaridis, and Marinka Zitnik.
\newblock Units: Building a unified time series model.
\newblock \emph{arXiv}, 2024.
\newblock URL \url{https://arxiv.org/pdf/2403.00131.pdf}.

\bibitem[Goswami et~al.(2024)Goswami, Szafer, Choudhry, Cai, Li, and Dubrawski]{goswami2024moment}
Mononito Goswami, Konrad Szafer, Arjun Choudhry, Yifu Cai, Shuo Li, and Artur Dubrawski.
\newblock Moment: A family of open time-series foundation models.
\newblock \emph{arXiv preprint arXiv:2402.03885}, 2024.

\bibitem[Guo et~al.(2023)Guo, Wang, Du, and Wang]{guo2023contranorm}
Xiaojun Guo, Yifei Wang, Tianqi Du, and Yisen Wang.
\newblock Contranorm: A contrastive learning perspective on oversmoothing and beyond.
\newblock In \emph{The Eleventh International Conference on Learning Representations}, 2023.

\bibitem[Hollmann et~al.(2025)Hollmann, M{\"u}ller, Purucker, Krishnakumar, K{\"o}rfer, Hoo, Schirrmeister, and Hutter]{hollmann2025accurate}
Noah Hollmann, Samuel M{\"u}ller, Lennart Purucker, Arjun Krishnakumar, Max K{\"o}rfer, Shi~Bin Hoo, Robin~Tibor Schirrmeister, and Frank Hutter.
\newblock Accurate predictions on small data with a tabular foundation model.
\newblock \emph{Nature}, 637\penalty0 (8045):\penalty0 319--326, 2025.

\bibitem[Hu et~al.(2022)Hu, Wallis, Allen-Zhu, Li, Wang, Wang, Chen, et~al.]{hulora}
Edward~J Hu, Phillip Wallis, Zeyuan Allen-Zhu, Yuanzhi Li, Shean Wang, Lu~Wang, Weizhu Chen, et~al.
\newblock Lora: Low-rank adaptation of large language models.
\newblock In \emph{International Conference on Learning Representations}, 2022.

\bibitem[Huet et~al.(2022)Huet, Navarro, and Rossi]{huet2022local}
Alexis Huet, Jose~Manuel Navarro, and Dario Rossi.
\newblock Local evaluation of time series anomaly detection algorithms.
\newblock In \emph{Proceedings of the ACM SIGKDD international conference on knowledge discovery \& data mining}, pp.\  635--645, 2022.

\bibitem[Hundman et~al.(2018)Hundman, Constantinou, Laporte, Colwell, and Soderstrom]{msl}
Kyle Hundman, Valentino Constantinou, Christopher Laporte, Ian Colwell, and Tom Soderstrom.
\newblock Detecting spacecraft anomalies using lstms and nonparametric dynamic thresholding.
\newblock In \emph{SIGKDD}, pp.\  387--395, 2018.

\bibitem[Kieu et~al.(2018)Kieu, Yang, and Jensen]{kieu2018outlier}
Tung Kieu, Bin Yang, and Christian~S Jensen.
\newblock Outlier detection for multidimensional time series using deep neural networks.
\newblock In \emph{2018 19th IEEE International Conference on Mobile Data Management (MDM)}, 2018.

\bibitem[Kim et~al.(2025)Kim, Mok, Lee, Lew, Kim, and Yoon]{kimcausality}
HyunGi Kim, Jisoo Mok, Dongjun Lee, Jaihyun Lew, Sungjae Kim, and Sungroh Yoon.
\newblock Causality-aware contrastive learning for robust multivariate time-series anomaly detection.
\newblock In \emph{Forty-second International Conference on Machine Learning}, 2025.

\bibitem[Li et~al.(2024{\natexlab{a}})Li, Huang, Wu, Liu, Yang, and Gui]{li2024hybrid}
Junxian Li, Keke Huang, Dehao Wu, Yishun Liu, Chunhua Yang, and Weihua Gui.
\newblock Hybrid variable dictionary learning for monitoring continuous and discrete variables in manufacturing processes.
\newblock \emph{Control Engineering Practice}, 149:\penalty0 105970, 2024{\natexlab{a}}.

\bibitem[Li et~al.(2025{\natexlab{a}})Li, Lu, Ren, and Kong]{li2025set}
Leyang Li, Shilin Lu, Yan Ren, and Adams Wai-Kin Kong.
\newblock Set you straight: Auto-steering denoising trajectories to sidestep unwanted concepts.
\newblock \emph{arXiv preprint arXiv:2504.12782}, 2025{\natexlab{a}}.

\bibitem[Li et~al.(2025{\natexlab{b}})Li, He, Zu, Li, Shi, Xie, and Zhang]{li2025multi}
Xiang Li, Yangfan He, Shuaishuai Zu, Zhengyang Li, Tianyu Shi, Yiting Xie, and Kevin Zhang.
\newblock Multi-modal large language model with rag strategies in soccer commentary generation.
\newblock In \emph{WACV}, pp.\  6197--6206, 2025{\natexlab{b}}.

\bibitem[Li et~al.(2025{\natexlab{c}})Li, Li, Song, Liu, Ji, Wang, Ma, Yu, Liu, Wang, et~al.]{li2025swea}
Xiaopeng Li, Shasha Li, Shezheng Song, Huijun Liu, Bin Ji, Xi~Wang, Jun Ma, Jie Yu, Xiaodong Liu, Jing Wang, et~al.
\newblock Swea: Updating factual knowledge in large language models via subject word embedding altering.
\newblock In \emph{Proceedings of the AAAI Conference on Artificial Intelligence}, volume~39, pp.\  24494--24502, 2025{\natexlab{c}}.

\bibitem[Li et~al.(2024{\natexlab{b}})Li, Wang, and Katsaggelos]{Li_2024_BMVC}
Yijie Li, Hewei Wang, and Aggelos Katsaggelos.
\newblock {CPDR: Towards Highly-Efficient Salient Object Detection via Crossed Post-decoder Refinement}.
\newblock In \emph{35th British Machine Vision Conference 2024, {BMVC} 2024, Glasgow, UK, November 25-28, 2024}, 2024{\natexlab{b}}.

\bibitem[Li et~al.(2025{\natexlab{d}})Li, Qiu, Chen, Wang, Cheng, Shu, Hu, Guo, Zhou, Jensen, and Yang]{li25TSFM-Bench}
Zhe Li, Xiangfei Qiu, Peng Chen, Yihang Wang, Hanyin Cheng, Yang Shu, Jilin Hu, Chenjuan Guo, Aoying Zhou, Christian~S. Jensen, and Bin Yang.
\newblock Tsfm-bench: A comprehensive and unified benchmark of foundation models for time series forecasting.
\newblock In \emph{SIGKDD}, pp.\  5595–5606, 2025{\natexlab{d}}.

\bibitem[Liu et~al.(2008)Liu, Ting, and Zhou]{liu2008isolation}
Fei~Tony Liu, Kai~Ming Ting, and Zhi-Hua Zhou.
\newblock Isolation forest.
\newblock In \emph{Proceedings of the 2008 IEEE international conference on data mining}, pp.\  413--422, 2008.

\bibitem[Liu \& Paparrizos(2024)Liu and Paparrizos]{liu2024elephant}
Qinghua Liu and John Paparrizos.
\newblock The elephant in the room: Towards a reliable time-series anomaly detection benchmark.
\newblock In \emph{NeurIPS}, 2024.

\bibitem[Liu et~al.(2024{\natexlab{a}})Liu, Boniol, Palpanas, and Paparrizos]{liu2024time}
Qinghua Liu, Paul Boniol, Themis Palpanas, and John Paparrizos.
\newblock Time-series anomaly detection: Overview and new trends.
\newblock \emph{Proc. {VLDB} Endow.}, 17\penalty0 (12):\penalty0 4229--4232, 2024{\natexlab{a}}.

\bibitem[Liu et~al.(2024{\natexlab{b}})Liu, Zhang, Li, Huang, Wang, and Long]{timer}
Yong Liu, Haoran Zhang, Chenyu Li, Xiangdong Huang, Jianmin Wang, and Mingsheng Long.
\newblock Timer: Generative pre-trained transformers are large time series models.
\newblock In \emph{International Conference on Machine Learning}, pp.\  32369--32399. PMLR, 2024{\natexlab{b}}.

\bibitem[Liu et~al.(2025)Liu, Qin, Shi, Chen, Yang, Huang, Wang, and Long]{liu2025sundial}
Yong Liu, Guo Qin, Zhiyuan Shi, Zhi Chen, Caiyin Yang, Xiangdong Huang, Jianmin Wang, and Mingsheng Long.
\newblock Sundial: A family of highly capable time series foundation models.
\newblock \emph{arXiv preprint arXiv:2502.00816}, 2025.

\bibitem[Lovie(2005)]{lovie2005coefficient}
Pat Lovie.
\newblock Coefficient of variation.
\newblock \emph{Encyclopedia of statistics in behavioral science}, 2005.

\bibitem[Lu et~al.(2023)Lu, Liu, and Kong]{lu2023tf}
Shilin Lu, Yanzhu Liu, and Adams Wai-Kin Kong.
\newblock Tf-icon: Diffusion-based training-free cross-domain image composition.
\newblock In \emph{ICCV}, pp.\  2294--2305, 2023.

\bibitem[Lu et~al.(2024{\natexlab{a}})Lu, Wang, Li, Liu, and Kong]{lu2024mace}
Shilin Lu, Zilan Wang, Leyang Li, Yanzhu Liu, and Adams Wai-Kin Kong.
\newblock Mace: Mass concept erasure in diffusion models.
\newblock In \emph{CVPR}, pp.\  6430--6440, 2024{\natexlab{a}}.

\bibitem[Lu et~al.(2024{\natexlab{b}})Lu, Zhou, Lu, Zhu, and Kong]{lu2024robust}
Shilin Lu, Zihan Zhou, Jiayou Lu, Yuanzhi Zhu, and Adams Wai-Kin Kong.
\newblock Robust watermarking using generative priors against image editing: From benchmarking to advances.
\newblock \emph{arXiv preprint arXiv:2410.18775}, 2024{\natexlab{b}}.

\bibitem[Luo \& Wang(2024)Luo and Wang]{luo2024moderntcn}
Donghao Luo and Xue Wang.
\newblock Moderntcn: A modern pure convolution structure for general time series analysis.
\newblock In \emph{The Twelfth International Conference on Learning Representations}, 2024.

\bibitem[Mathur \& Tippenhauer(2016)Mathur and Tippenhauer]{swat}
Aditya~P Mathur and Nils~Ole Tippenhauer.
\newblock Swat: A water treatment testbed for research and training on ics security.
\newblock In \emph{CySWater}, pp.\  31--36, 2016.

\bibitem[Paparrizos et~al.(2022)Paparrizos, Boniol, Palpanas, Tsay, Elmore, and Franklin]{paparrizos2022volume}
John Paparrizos, Paul Boniol, Themis Palpanas, Ruey~S Tsay, Aaron Elmore, and Michael~J Franklin.
\newblock Volume under the surface: a new accuracy evaluation measure for time-series anomaly detection.
\newblock \emph{Proceedings of the VLDB Endowment}, 15\penalty0 (11):\penalty0 2774--2787, 2022.

\bibitem[Qiao et~al.(2025)Qiao, Liu, Zhang, Jin, Pham, Wen, Suganthan, Jiang, and Ramasamy]{qiao2025multi}
Zhongzheng Qiao, Chenghao Liu, Yiming Zhang, Ming Jin, Quang Pham, Qingsong Wen, PN~Suganthan, Xudong Jiang, and Savitha Ramasamy.
\newblock Multi-scale finetuning for encoder-based time series foundation models.
\newblock \emph{arXiv preprint arXiv:2506.14087}, 2025.

\bibitem[Qiu et~al.(2024)Qiu, Hu, Zhou, Wu, Du, Zhang, Guo, Zhou, Jensen, Sheng, and Yang]{qiu2024tfb}
Xiangfei Qiu, Jilin Hu, Lekui Zhou, Xingjian Wu, Junyang Du, Buang Zhang, Chenjuan Guo, Aoying Zhou, Christian~S. Jensen, Zhenli Sheng, and Bin Yang.
\newblock {TFB}: Towards comprehensive and fair benchmarking of time series forecasting methods.
\newblock In \emph{Proc. {VLDB} Endow.}, pp.\  2363--2377, 2024.

\bibitem[Qiu et~al.(2025{\natexlab{a}})Qiu, Li, Pang, Pan, Wu, Yang, Hu, Shu, Lu, Yang, Guo, Zhou, Jensen, and Yang]{qiu2025easytime}
Xiangfei Qiu, Xiuwen Li, Ruiyang Pang, Zhicheng Pan, Xingjian Wu, Liu Yang, Jilin Hu, Yang Shu, Xuesong Lu, Chengcheng Yang, Chenjuan Guo, Aoying Zhou, Christian~S. Jensen, and Bin Yang.
\newblock {EasyTime}: Time series forecasting made easy.
\newblock In \emph{ICDE}, 2025{\natexlab{a}}.

\bibitem[Qiu et~al.(2025{\natexlab{b}})Qiu, Li, Qiu, Hu, Zhou, Wu, Li, Guo, Zhou, Sheng, Hu, Jensen, and Yang]{qiu2025tab}
Xiangfei Qiu, Zhe Li, Wanghui Qiu, Shiyan Hu, Lekui Zhou, Xingjian Wu, Zhengyu Li, Chenjuan Guo, Aoying Zhou, Zhenli Sheng, Jilin Hu, Christian~S. Jensen, and Bin Yang.
\newblock Tab: Unified benchmarking of time series anomaly detection methods.
\newblock In \emph{Proc. {VLDB} Endow.}, pp.\  2775--2789, 2025{\natexlab{b}}.

\bibitem[Qiu et~al.(2025{\natexlab{c}})Qiu, Wu, Lin, Guo, Hu, and Yang]{qiu2025duet}
Xiangfei Qiu, Xingjian Wu, Yan Lin, Chenjuan Guo, Jilin Hu, and Bin Yang.
\newblock {DUET}: Dual clustering enhanced multivariate time series forecasting.
\newblock In \emph{SIGKDD}, pp.\  1185--1196, 2025{\natexlab{c}}.

\bibitem[Ramaswamy et~al.(2000)Ramaswamy, Rastogi, and Shim]{ramaswamy2000efficient}
Sridhar Ramaswamy, Rajeev Rastogi, and Kyuseok Shim.
\newblock Efficient algorithms for mining outliers from large data sets.
\newblock In \emph{Proceedings of the 2000 ACM SIGMOD international conference on management of data}, pp.\  427--438, 2000.

\bibitem[Shan et~al.(2016)Shan, Hoens, Jiao, Wang, Yu, and Mao]{shan2016deep}
Ying Shan, T~Ryan Hoens, Jian Jiao, Haijing Wang, Dong Yu, and JC~Mao.
\newblock Deep crossing: Web-scale modeling without manually crafted combinatorial features.
\newblock In \emph{Proceedings of the 22nd ACM SIGKDD international conference on knowledge discovery and data mining}, pp.\  255--262, 2016.

\bibitem[Shentu et~al.(2024)Shentu, Li, Zhao, Shu, Rao, Pan, Yang, and Guo]{shentu2024towards}
Qichao Shentu, Beibu Li, Kai Zhao, Yang Shu, Zhongwen Rao, Lujia Pan, Bin Yang, and Chenjuan Guo.
\newblock Towards a general time series anomaly detector with adaptive bottlenecks and dual adversarial decoders.
\newblock \emph{arXiv preprint arXiv:2405.15273}, 2024.

\bibitem[Shi et~al.(2024)Shi, Wang, Nie, Li, Ye, Wen, and Jin]{timemoe}
Xiaoming Shi, Shiyu Wang, Yuqi Nie, Dianqi Li, Zhou Ye, Qingsong Wen, and Ming Jin.
\newblock Time-moe: Billion-scale time series foundation models with mixture of experts.
\newblock \emph{arXiv e-prints}, pp.\  arXiv--2409, 2024.

\bibitem[Stewart(1993)]{stewart1993early}
Gilbert~W Stewart.
\newblock On the early history of the singular value decomposition.
\newblock \emph{SIAM review}, 35\penalty0 (4):\penalty0 551--566, 1993.

\bibitem[Tatbul et~al.(2018)Tatbul, Lee, Zdonik, Alam, and Gottschlich]{tatbul2018precision}
Nesime Tatbul, Tae~Jun Lee, Stan Zdonik, Mejbah Alam, and Justin Gottschlich.
\newblock Precision and recall for time series.
\newblock \emph{Advances in neural information processing systems}, 31, 2018.

\bibitem[Tuli et~al.(2022)Tuli, Casale, and Jennings]{tuli2022tranad}
Shreshth Tuli, Giuliano Casale, and Nicholas~R Jennings.
\newblock Tranad: Deep transformer networks for anomaly detection in multivariate time series data.
\newblock \emph{arXiv preprint arXiv:2201.07284}, 2022.

\bibitem[Vaswani et~al.(2017)Vaswani, Shazeer, Parmar, Uszkoreit, Jones, Gomez, Kaiser, and Polosukhin]{DBLP:conf/nips/VaswaniSPUJGKP17}
Ashish Vaswani, Noam Shazeer, Niki Parmar, Jakob Uszkoreit, Llion Jones, Aidan~N. Gomez, Lukasz Kaiser, and Illia Polosukhin.
\newblock Attention is all you need.
\newblock In \emph{Neural Information Processing Systems NeurIPS}, pp.\  5998--6008, 2017.

\bibitem[von Birgelen \& Niggemann(2018)von Birgelen and Niggemann]{Genesis}
Alexander von Birgelen and Oliver Niggemann.
\newblock Anomaly detection and localization for cyber-physical production systems with self-organizing maps.
\newblock \emph{IMPROVE-Innovative Modelling Approaches for Production Systems to Raise Validatable Efficiency: Intelligent Methods for the Factory of the Future}, pp.\  55--71, 2018.

\bibitem[Wang et~al.(2025{\natexlab{a}})Wang, He, Sun, Li, and Shi]{wang2025unitmge}
Ruoyu Wang, Yangfan He, Tengjiao Sun, Xiang Li, and Tianyu Shi.
\newblock Unitmge: Uniform text-motion generation and editing model via diffusion.
\newblock In \emph{WACV}, pp.\  6104--6114, 2025{\natexlab{a}}.

\bibitem[Wang et~al.(2024)Wang, Qiu, Chen, Zhao, Shu, Rao, Pan, Yang, and Guo]{wang2024rose}
Yihang Wang, Yuying Qiu, Peng Chen, Kai Zhao, Yang Shu, Zhongwen Rao, Lujia Pan, Bin Yang, and Chenjuan Guo.
\newblock {ROSE}: Register assisted general time series forecasting with decomposed frequency learning.
\newblock \emph{arXiv preprint arXiv:2405.17478}, 2024.

\bibitem[Wang et~al.(2025{\natexlab{b}})Wang, Qiu, Chen, Shu, Rao, Pan, Yang, and Guo]{wang2025lightgts}
Yihang Wang, Yuying Qiu, Peng Chen, Yang Shu, Zhongwen Rao, Lujia Pan, Bin Yang, and Chenjuan Guo.
\newblock Lightgts: A lightweight general time series forecasting model.
\newblock \emph{arXiv preprint arXiv:2506.06005}, 2025{\natexlab{b}}.

\bibitem[Wen et~al.(2022)Wen, Yang, Zhou, and Sun]{wen2022robust}
Qingsong Wen, Linxiao Yang, Tian Zhou, and Liang Sun.
\newblock Robust time series analysis and applications: An industrial perspective.
\newblock In \emph{Proceedings of the 28th ACM SIGKDD Conference on Knowledge Discovery and Data Mining}, pp.\  4836--4837, 2022.

\bibitem[Wu et~al.(2023)Wu, Hu, Liu, Zhou, Wang, and Long]{wu2022timesnet}
Haixu Wu, Tengge Hu, Yong Liu, Hang Zhou, Jianmin Wang, and Mingsheng Long.
\newblock Timesnet: Temporal 2d-variation modeling for general time series analysis.
\newblock In \emph{ICLR}, 2023.

\bibitem[Wu et~al.(2025{\natexlab{a}})Wu, Qiu, Gao, Hu, Yang, and Guo]{wu2025k2vae}
Xingjian Wu, Xiangfei Qiu, Hongfan Gao, Jilin Hu, Bin Yang, and Chenjuan Guo.
\newblock {K${}^2$VAE}: A koopman-kalman enhanced variational autoencoder for probabilistic time series forecasting.
\newblock In \emph{ICML}, 2025{\natexlab{a}}.

\bibitem[Wu et~al.(2025{\natexlab{b}})Wu, Qiu, Li, Wang, Hu, Guo, Xiong, and Yang]{wu2024catch}
Xingjian Wu, Xiangfei Qiu, Zhengyu Li, Yihang Wang, Jilin Hu, Chenjuan Guo, Hui Xiong, and Bin Yang.
\newblock {CATCH}: Channel-aware multivariate time series anomaly detection via frequency patching.
\newblock In \emph{ICLR}, 2025{\natexlab{b}}.

\bibitem[Wu et~al.(2025{\natexlab{c}})Wu, Qiu, Li, Wang, Hu, Guo, Xiong, and Yang]{wucatch}
Xingjian Wu, Xiangfei Qiu, Zhengyu Li, Yihang Wang, Jilin Hu, Chenjuan Guo, Hui Xiong, and Bin Yang.
\newblock Catch: Channel-aware multivariate time series anomaly detection via frequency patching.
\newblock In \emph{International Conference on Learning Representations (ICLR)}, 2025{\natexlab{c}}.

\bibitem[Wu et~al.(2024)Wu, Wu, Yang, Zhou, Guo, Qiu, Hu, Sheng, and Jensen]{AutoCTS++}
Xinle Wu, Xingjian Wu, Bin Yang, Lekui Zhou, Chenjuan Guo, Xiangfei Qiu, Jilin Hu, Zhenli Sheng, and Christian~S. Jensen.
\newblock {AutoCTS++}: zero-shot joint neural architecture and hyperparameter search for correlated time series forecasting.
\newblock \emph{{VLDB} J.}, 33\penalty0 (5):\penalty0 1743--1770, 2024.

\bibitem[Xie et~al.(2025)Xie, Zhang, and Babar]{xie2025multivariate}
Yongzheng Xie, Hongyu Zhang, and Muhammad~Ali Babar.
\newblock Multivariate time series anomaly detection by capturing coarse-grained intra-and inter-variate dependencies.
\newblock In \emph{Proceedings of the ACM on Web Conference 2025}, pp.\  697--705, 2025.

\bibitem[Xu et~al.(2021)Xu, Wu, Wang, and Long]{xu2021anomaly}
Jiehui Xu, Haixu Wu, Jianmin Wang, and Mingsheng Long.
\newblock Anomaly {Transformer}: Time series anomaly detection with association discrepancy.
\newblock In \emph{International Conference on Learning Representations}, 2021.

\bibitem[Xu et~al.(2025)Xu, Cheng, Guo, Gao, Hu, Yang, and Yang]{xu2024mmpath}
Ronghui Xu, Hanyin Cheng, Chenjuan Guo, Hongfan Gao, Jilin Hu, Sean~Bin Yang, and Bin Yang.
\newblock Mm-path: Multi-modal, multi-granularity path representation learning.
\newblock In \emph{SIGKDD}, pp.\  1703–1714, 2025.

\bibitem[Yang et~al.(2024)Yang, He, Tian, Chen, Wang, Shi, Heydarian, and Liu]{yang2024wcdt}
Chen Yang, Yangfan He, Aaron~Xuxiang Tian, Dong Chen, Jianhui Wang, Tianyu Shi, Arsalan Heydarian, and Pei Liu.
\newblock Wcdt: World-centric diffusion transformer for traffic scene generation.
\newblock \emph{arXiv preprint arXiv:2404.02082}, 2024.

\bibitem[Yang et~al.(2023{\natexlab{a}})Yang, Li, Shi, Li, Hu, Li, and Yuan]{yang2023sgdp}
Yiyuan Yang, Rongshang Li, Qiquan Shi, Xijun Li, Gang Hu, Xing Li, and Mingxuan Yuan.
\newblock Sgdp: A stream-graph neural network based data prefetcher.
\newblock In \emph{2023 International Joint Conference on Neural Networks (IJCNN)}, pp.\  1--8, 2023{\natexlab{a}}.

\bibitem[Yang et~al.(2023{\natexlab{b}})Yang, Zhang, Zhou, Wen, and Sun]{yang2023dcdetector}
Yiyuan Yang, Chaoli Zhang, Tian Zhou, Qingsong Wen, and Liang Sun.
\newblock Dcdetector: Dual attention contrastive representation learning for time series anomaly detection.
\newblock In \emph{Proceedings of the 29th ACM SIGKDD Conference on Knowledge Discovery and Data Mining}, pp.\  3033--3045, 2023{\natexlab{b}}.

\bibitem[Yoon et~al.(2020)Yoon, Lee, and Lee]{yoon2020ultrafast}
Susik Yoon, Jae-Gil Lee, and Byung~Suk Lee.
\newblock Ultrafast local outlier detection from a data stream with stationary region skipping.
\newblock In \emph{SIGKDD}, pp.\  1181--1191, 2020.

\bibitem[Zhong et~al.(2025{\natexlab{a}})Zhong, Yu, Xi, Xu, Cao, Yang, Yang, and You]{simad}
Zhijie Zhong, Zhiwen Yu, Xing Xi, Yue Xu, Wenming Cao, Yiyuan Yang, Kaixiang Yang, and Jane You.
\newblock Simad: A simple dissimilarity-based approach for time-series anomaly detection.
\newblock \emph{IEEE Transactions on Neural Networks and Learning Systems}, 2025{\natexlab{a}}.

\bibitem[Zhong et~al.(2025{\natexlab{b}})Zhong, Yu, Xi, Xu, Cao, Yang, Yang, and You]{zhong2025simad}
Zhijie Zhong, Zhiwen Yu, Xing Xi, Yue Xu, Wenming Cao, Yiyuan Yang, Kaixiang Yang, and Jane You.
\newblock Simad: A simple dissimilarity-based approach for time-series anomaly detection.
\newblock \emph{IEEE Transactions on Neural Networks and Learning Systems}, 2025{\natexlab{b}}.

\bibitem[Zhou et~al.(2022)Zhou, Lei, Liu, Du, Huang, Zhao, Dai, Le, Laudon, et~al.]{zhou2022mixture}
Yanqi Zhou, Tao Lei, Hanxiao Liu, Nan Du, Yanping Huang, Vincent Zhao, Andrew~M Dai, Quoc~V Le, James Laudon, et~al.
\newblock Mixture-of-experts with expert choice routing.
\newblock \emph{Advances in Neural Information Processing Systems}, 35:\penalty0 7103--7114, 2022.

\bibitem[Zhou et~al.(2025)Zhou, He, Su, Han, Jang, Bertasius, Bansal, and Yao]{zhou2025reagent}
Yiyang Zhou, Yangfan He, Yaofeng Su, Siwei Han, Joel Jang, Gedas Bertasius, Mohit Bansal, and Huaxiu Yao.
\newblock Reagent-v: A reward-driven multi-agent framework for video understanding.
\newblock \emph{arXiv preprint arXiv:2506.01300}, 2025.

\end{thebibliography}
\bibliographystyle{iclr2026-conference}

\clearpage
\appendix
\appendix

\section{Experimental Details}
\label{app: exp details}
\subsection{Datasets}
\label{app: datasets}

% 我们选择了5个经典的真实世界多变量异常检测数据集来评估我们的方法，包括：
To evaluate our method, we selected five classic real-world multivariate anomaly detection datasets:
1) \textbf{MSL}~\citep{msl}: Spacecraft incident and anomaly data from the MSL Curiosity rover. 
2) \textbf{SMAP}~\citep{msl}:  Presents the soil samples and telemetry information used by the Mars rover. 
3) \textbf{SWAT}~\citep{swat}:  Obtained from 51 sensors of the critical infrastructure system under continuous operations. 
4) \textbf{Genesis}~\citep{Genesis}: A sensor and control signals dataset collected from cyber-physical production systems. 
5) \textbf{NYC}~\citep{nyc}: The Transportation dataset provides information on taxi and ride-hailing trips in New York. 
% 其中的每个数据集都包含了大量的Stae variables, 具体的统计指标如表1所示。
Each of these datasets contains a significant number of \textit{state variables}, with detailed statistics presented in Table ~\ref{Multivariate datasets}.
\begin{table}[h]
\caption{Statistics of multivariate datasets (\footnotesize{AR: anomaly ratio}).}
\label{Multivariate datasets}
\centering
\resizebox{0.8\columnwidth}{!}{
\centering
\begin{tabular}{@{}llccccc@{}}
\toprule
\textbf{Dataset}      & \textbf{Domain}   & 
 \textbf{\begin{tabular}[c]{@{}c@{}}Numeral\\ Numbers\end{tabular}}& 
  \textbf{\begin{tabular}[c]{@{}c@{}}State\\ Numbers\end{tabular}}& 
 \textbf{AR~(\%)}&
 \textbf{\begin{tabular}[c]{@{}c@{}}Avg Total\\ Length\end{tabular}}& 
 \textbf{\begin{tabular}[c]{@{}c@{}}Avg Test\\ Length\end{tabular}} \\ 
 \midrule
MSL     & Spacecraft  & 1 & 54  & 5.88    & 132,046            & 73,729  \\
SMAP&  Spacecraft  &1 & 24&  9.72&  562,800&  427,617\\
SWAT & Water treatment &28&23 & 5.78 &944,919& 449,919 \\
Genesis        & Machinery &5  & 13     & 0.31   & 16,220          & 12,616 \\
NYC     & Transport & 1  & 2  & 0.57    & 17,520            & 4,416\\ 
\bottomrule
\end{tabular}
}
\end{table}

\subsection{Time Series Foundation Models for Anomaly Detection}
\label{app: backbones}

% 近年来深度学习在图像，文本，时空等任务中取得了广泛的成功。
In recent years, deep learning~\citep{wu2025k2vae,qiu2025duet,AutoCTS++} has achieved widespread success in tasks such as image~\citep{lu2024mace,lu2023tf,lu2024robust,zhou2025reagent}, language~\citep{li2025multi,wang2025unitmge}, and spatio-temporal analysis~\citep{xu2024mmpath,li2025set,yang2024wcdt,wang2025unitmge,qiu2025easytime,wu2024catch}.

In the realm of time series analysis, numerous models have surfaced in recent years~\citep{qiu2025easytime,wu2024catch,qiu2024tfb,li25TSFM-Bench}. We choose models which designed for multi-task or specialised for anomaly detection, including task-specific model DADA~\citep{shentu2024towards} and task-general models: UniTS~\citep{units}, Moment~\citep{goswami2024moment}, Timer~\citep{timer}. The specific descriptions for each of these models are listed in Table~\ref{Descriptions of baselines.}.

\begin{table*}[h!]
\centering
\caption{Descriptions of Time Series Foundation Models for Anomaly Detection in experiments.}
\label{Descriptions of baselines.}
  \resizebox{0.8\linewidth}{!}{
    \begin{tabular}{l|l}
     \toprule
        \textbf{Models} & \textbf{Descriptions} \\ \midrule

        DADA & \makecell[l]{DADA is a general time series anomaly detector pre-trained on multi-domain\\ data, utilizing adaptive bottlenecks and dual adversarial decoders. It is designed \\to detect by learning to differentiate between normal and abnormal patterns.}\\\midrule
        
        UniTS & \makecell[l]{UNITS is a unified, multi-task transformer model that uses task tokenization \\ to handle a variety of time series tasks like forecasting, classification,\\ imputation, and anomaly detection within a single framework.}\\\midrule
        
        Moment & \makecell[l]{Moment is a transformer system pre-trained on a masked time series task. \\It reconstructs masked portions of time series for tasks like forecasting,\\ classification, anomaly detection, and imputation.} \\\midrule
       
        Timer & \makecell[l]{Timer is a GPT-style autoregressive model for time series analysis, \\predicting the next token in single-series sequences. It supports tasks like \\forecasting, imputation, and anomaly detection across different time series.}\\ \midrule
    \end{tabular}}
\end{table*}

\subsection{Metrics for Time Series Anomaly Detection}
\label{app: Metrics}

The metrics used for evaluation can be divided into two categories: Threshold-relevant and Threshold-irrelevant. Threshold-relevant metrics depend on threshold parameters to convert anomaly scores into anomaly labels. Among them, Threshold-irrelevant metrics, on the other hand, evaluate the performance of the model based on the raw anomaly scores it generates. Please refer to Table~\ref{Overview of evaluation metrics} for an overview of these metrics. 
\begin{table}[h]
\caption{Overview of evaluation metrics.}
\label{Overview of evaluation metrics}
\centering
\resizebox{1\linewidth}{!}{
\begin{tabular}{@{}c|l|l|l@{}}
\toprule
\multicolumn{1}{l|}{\textbf{Category}} & \textbf{Metric} & \textbf{Abbreviation} & \textbf{Short summary} \\ \midrule
\multirow{7}{*}{\rot{\makecell{\textbf{Threshold-}\\\textbf{relevant}}}} & Accuracy & Acc & Measures the proportion of correct predictions among the total number of predictions.\\\cmidrule{2-4}
 & Range-Precision & R-P~\cite{tatbul2018precision} & These metrics' definitions consider several factors: the ratio of detected anomaly subsequences to the total number of anomalies,  \\
 & Range-Recall & R-R~\cite{tatbul2018precision} &  the ratio of detected point outliers to total point outliers, the relative position of true positives within each anomaly subsequence,\\
 & Range-F1-score & R-F1~\cite{tatbul2018precision} & and the number of fragmented prediction regions corresponding to one real anomaly subsequence.\\ \cmidrule{2-4}
 & Affiliated-Precision & Aff-P~\cite{huet2022local} & These metrics' definitions are the extension of the classical precision/recall/f1-score for time series anomaly detection that is  \\
 & Affiliated-Recall & Aff-R~\cite{huet2022local} & local (each ground truth event is considered separately), parameter-free, and applicable generically on both point and subsequence  \\
 & Affiliated-F1-score & Aff-F1~\cite{huet2022local} & anomalies. Besides the construction of these metrics makes them both theoretically principled and practically useful.\\\cmidrule{1-4}
\multirow{6}{*}{\rot{\makecell{\textbf{Threshold-}\\\textbf{irrelevant}}}} & AUC-PR & A-P~\cite{davis2006relationship} & Measures the area under the curve corresponding to Recall on the x-axis and Precision on the y-axis at various threshold settings. \\\cmidrule{2-4}
 & AUC-ROC & A-R~\cite{fawcett2006introduction} & Measures the area under the curve
corresponding to FPR on the x-axis and TPR on the y-axis at various threshold settings.  \\  \cmidrule{2-4}
 & R-AUC-PR & R-A-P~\cite{paparrizos2022volume} & They mitigate the issue that AUC-PR and AUC-ROC are designed for point-based anomaly detection, where each point is\\ 
 & R-AUC-ROC & R-A-R~\cite{paparrizos2022volume} &    assigned equal weight in calculating the overall AUC. This makes them unsuitable for evaluating subsequence anomalies.\\ \cmidrule{2-4}
 & VUS-PR & V-PR~\cite{paparrizos2022volume} &  VUS is an extension of the ROC and PR curves. It introduces a buffer region at the outliers’ boundaries, thereby accommodating \\ 
 & VUS-ROC & V-ROC~\cite{paparrizos2022volume} &  the false tolerance of labeling in the ground  truth and assigning higher anomaly scores near the outlier boundaries.\\ \bottomrule
\end{tabular}}
\vspace{-10mm}
\end{table}

\section{More Analysis on STAR}

\subsection{Applicability for dataset}
\label{app: Applicability for dataset}
\begin{wraptable}{r}{0.6\columnwidth}
\vspace{-2.5mm}
\centering
\caption{Evaluate STAR on datasets containing solely numerical variables.}
\label{tab: generalization}
\begin{threeparttable}
\resizebox{0.6\columnwidth}{!}{
\begin{tabular}{l|ccc|ccc|ccc}
        \specialrule{0.8pt}{0pt}{2pt}
        \textbf{Dataset} & \multicolumn{3}{c|}{\textbf{PSM}} & \multicolumn{3}{c|}{\textbf{PUMP}}& \multicolumn{3}{c}{\textbf{ECG}} \\ \cmidrule{1-10}
        \textbf{Metric} & \textbf{A-R} & \textbf{V-R} & \textbf{V-P} & \textbf{A-R} & \textbf{V-R} & \textbf{V-P} & \textbf{A-R} & \textbf{V-R} & \textbf{V-P}  \\ \specialrule{0.8pt}{2pt}{2pt}
        \textbf{DADA} & 0.606  & 0.581  & 0.403  & 0.775  & 0.778  & 0.226 & 0.839  & 0.807  & 0.508\\ \cmidrule{1-10} 
        \textbf{DADA*} & 0.621          & 0.588          & 0.391          & 0.764          & 0.777          & 0.223          & 0.828          & 0.801          & 0.495\\ \cmidrule{1-10} 
        \textcolor{redorange}{\textbf{+STAR*}} & \textbf{0.636} & \textbf{0.592} & \textbf{0.412} & \textbf{0.807} & \textbf{0.807} & \textbf{0.236} & \textbf{0.844} & \textbf{0.811} & \textbf{0.514} \\ \specialrule{0.8pt}{2pt}{2pt} 
        \textbf{UniTS} & 0.582          & 0.573          & 0.377          & 0.415          & 0.573          & 0.211          & 0.872          & 0.840          & 0.522\\ \cmidrule{1-10} 
        \textbf{UniTS*}& 0.548          & 0.544          & 0.354          & 0.384          & 0.536          & 0.195          & 0.863          & 0.835          & 0.516\\ \cmidrule{1-10} 
        \textcolor{redorange}{\textbf{+STAR*}} &\textbf{0.607} & \textbf{0.602} & \textbf{0.401} & \textbf{0.436} & \textbf{0.598} & \textbf{0.217} & \textbf{0.884} & \textbf{0.848} & \textbf{0.537}\\  \specialrule{0.8pt}{2pt}{0pt}
    \end{tabular}}
\end{threeparttable}
\vspace{-2mm}
\end{wraptable}
% 虽然大部分的工业场景下，会存在各样的sate变量，但是也有一些特殊的或者受限的场景下，我们无法得到离散变量。
While most industrial scenarios feature a variety of \textit{state variables}, there are also special or constrained settings where such variables are unavailable. 
% 为了进一步拓展STAR的应用，我们尝试了主动去构造伪协变量作为\textit{state variables}交由STAR处理。
To further extend the applicability of STAR, we explored the approach of proactively constructing pseudo-covariates to serve as \textit{state variables} for processing by the STAR module. 
% 具体来说，模型数值变量本身就会带有状态信息，如温度，会代表系统处于高温还是低温状态，PH值包含了系统处于酸性环境还是碱性环境的前提。我们将这些数值变量划分为离散的状态变量，来构造伪协变量。
Specifically, numerical variables often inherently carry state information. For instance, temperature indicates whether the system is in a high or low-temperature state, while the pH value reflects the precondition of an acidic or alkaline environment. We leverage this property by discretizing these numerical variables into discrete states to construct our pseudo-covariates. We conducted this experiment on the PSM~\cite{abdulaal2021practical}  , PUMP~\cite{feng2021time}, and ECG~\cite{yoon2020ultrafast}  datasets, and the results are presented in Table~\ref{tab: generalization}. Rows marked with an asterisk (*) denote the use of pseudo-covariates. The results indicate that in most cases, the incorporation of \textit{state variables} still leads to a decrease in performance. Although the pseudo-covariates lack a well-defined physical meaning, STAR is nevertheless able to improve the performance of TSFMs to a certain degree.

\subsection{Visualization of Identity-guided State Encoder}
\label{app: Visualization}
\begin{figure*}[!h]
    \centering
\includegraphics[width=0.7\linewidth]{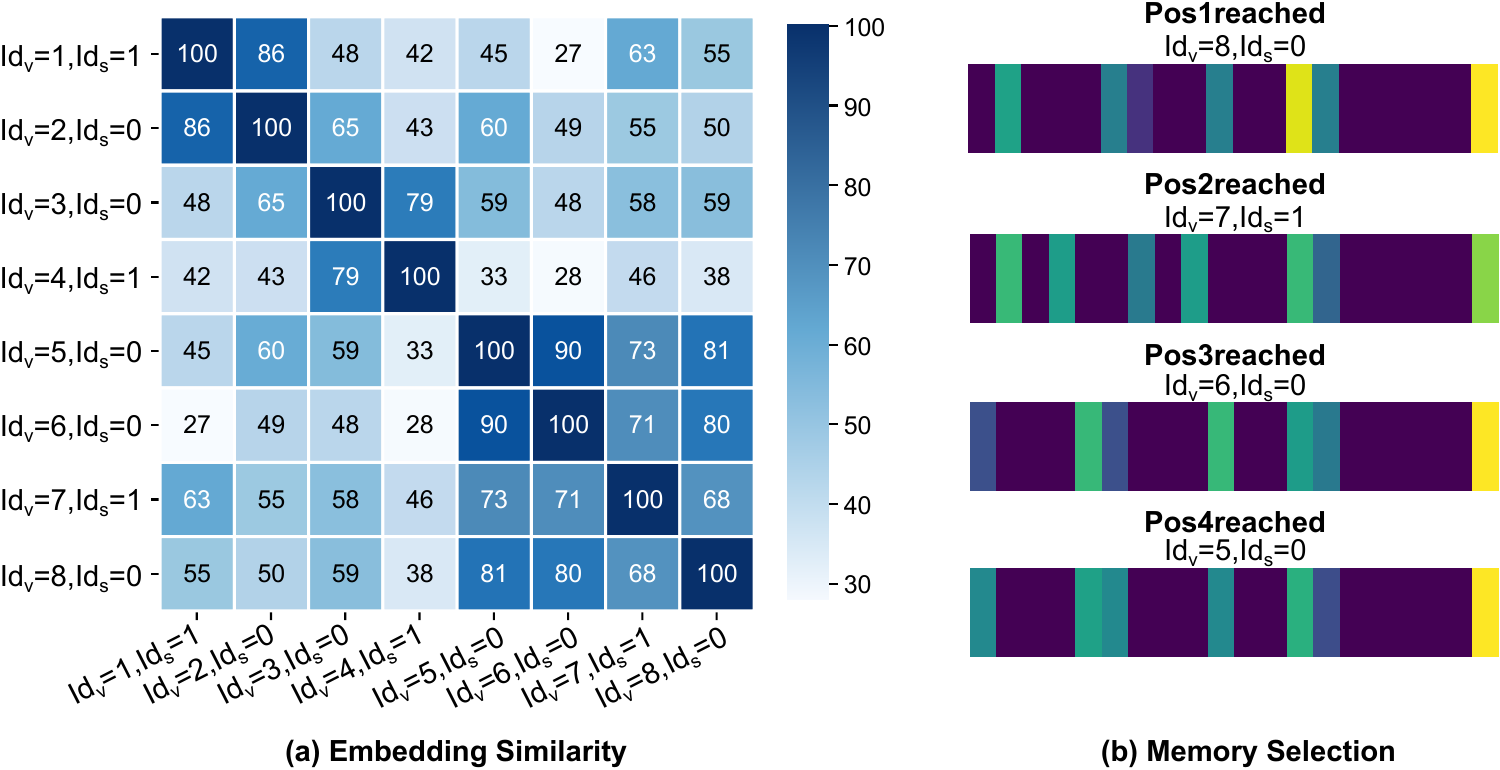}
    \caption{(a) shows the similarity between the embeddings output from \textit{Identity-guided State Encoder}. (b) shows the weight of selection in \textit{Memory Router}.}
\label{fig: visualization}
\vspace{-3mm}
\end{figure*}

% 为了进一步验证\textit{Identity-guided State Encoder}的有效性，我们对其输出的表征以及Memory Router的选择进行了可视化。
To further validate the effectiveness of the \textit{Identity-guided State Encoder}, we visualized its output embeddings as well as the selections made by the \textit{Memory Router}. 
% 我们选择了Genesis作为分析的数据集，其中的部分变量表述列于表1中。
We selected the Genesis~\citep{Genesis} for our analysis, and a subset of its variables is described in Table~\ref{app: Visualization}. 
\begin{wraptable}{r}{0.5\columnwidth}
\vspace{-2.5mm}
\centering
\caption{Evaluate STAR on datasets containing solely numerical variables.}
\label{tab: visualization}
\begin{threeparttable}
\resizebox{0.5\columnwidth}{!}{
\begin{tabular}{c | c | c | c}
        \specialrule{0.8pt}{0pt}{2pt}
        $\boldsymbol{\text{\textbf{Id}}_\text{v}}$& \textbf{Group} & \textbf{Variable name} & Descriptions \\ \specialrule{0.8pt}{2pt}{2pt}
        1& \multirow{2}{*}{Silder State} & Slider\_OUT & The slider is on the outside. \\
        2& & Slider\_IN & The slider is on the inside. \\\cmidrule{1-4} 
        3& \multirow{2}{*}{Detection State} & NonMetall & Non-metal detected. \\
        4& & Metall & Metal detected.\\\cmidrule{1-4} 
        5& \multirow{4}{*}{Position State} & Pos4reached & Arrived at Position 4. \\
        6& & Pos3reached & Arrived at Position 3. \\
        7& & Pos2reached & Arrived at Position 2. \\
        8& & Pos1reached & Arrived at Position 1. \\ \specialrule{0.8pt}{2pt}{0pt}
    \end{tabular}
    }
\end{threeparttable}
\vspace{-2mm}
\end{wraptable}
% 我们使用id_v表示变量identy，id_s表示表示state identy
We use $\text{Id}_\text{v}$ to denote the variable identity (index of \textit{state variable}), and $\text{Id}_\text{s}$ to denote the state identity (discrete value of \textit{state variable}).
% 图4展示了不同变量间embeddings的相似度。其中可以看到较为明显的聚类现象。具有相似语义信息的变量具有更高的相似度。
Figure~\ref{fig: visualization}a illustrates the similarity between embeddings of different variables, where a distinct clustering pattern is observed. This indicates that variables with similar semantic information exhibit higher similarity.
% 由于我们的router同时基于a,b，因此具有相同a或b的变量embedding会更加相似，但整体来看a的影响大于b。
Since our router is conditioned on both variable identity and state identity, embeddings for variables that share the same variable identity or state identity exhibit greater similarity. However, the overall influence of variable identity is more pronounced than that of state identity.
% 以上的观察也可以在图b中得到应证
These observations are further corroborated by Figure~\ref{fig: visualization}b.
\newpage

\subsection{Case study}

\label{app: case study}

\begin{figure*}[!h]
    \centering
    {
\includegraphics[width=0.9\linewidth]{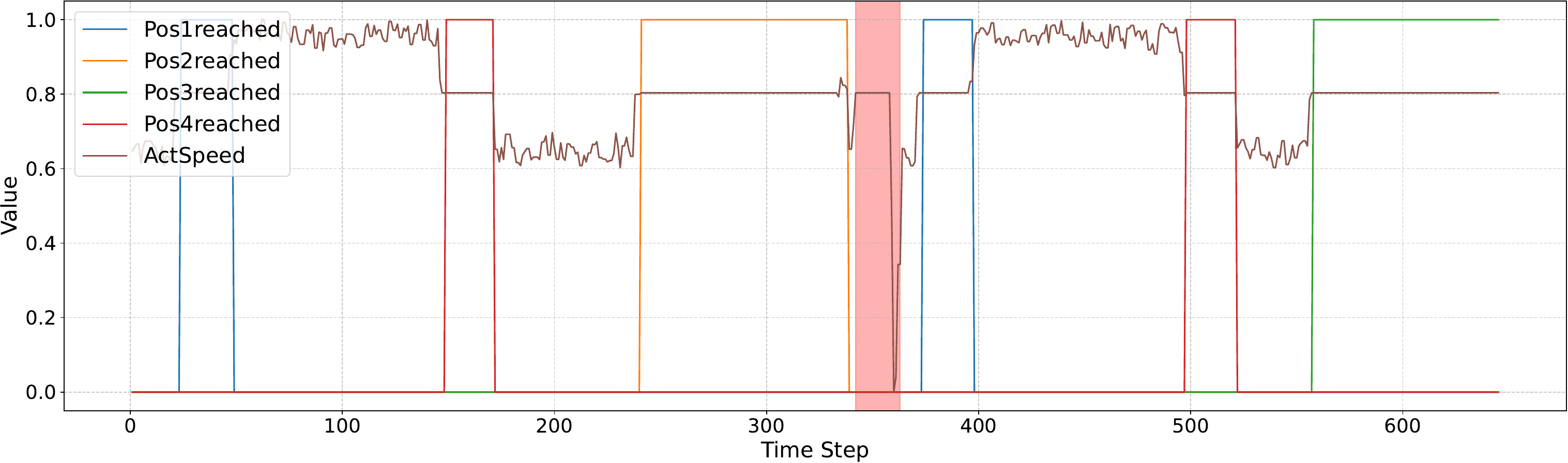}
}
    \caption{Case series in Genesis.}
\label{fig: case Genesis}
\vspace{-3mm}
\end{figure*}

\begin{figure*}[!h]
  \centering
  \subfloat[Window of case]
  {\includegraphics[width=0.3\textwidth]{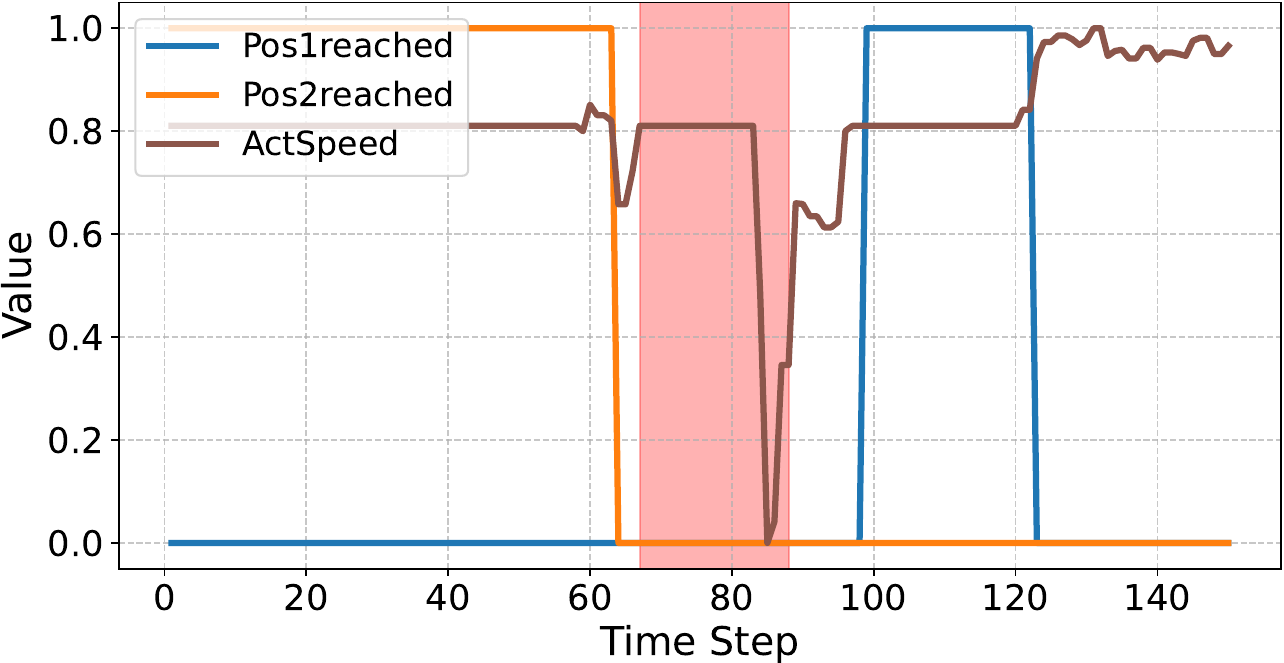}}  
  \hspace{3mm}
  \subfloat[Detect by DADA]
  {\includegraphics[width=0.3\textwidth]{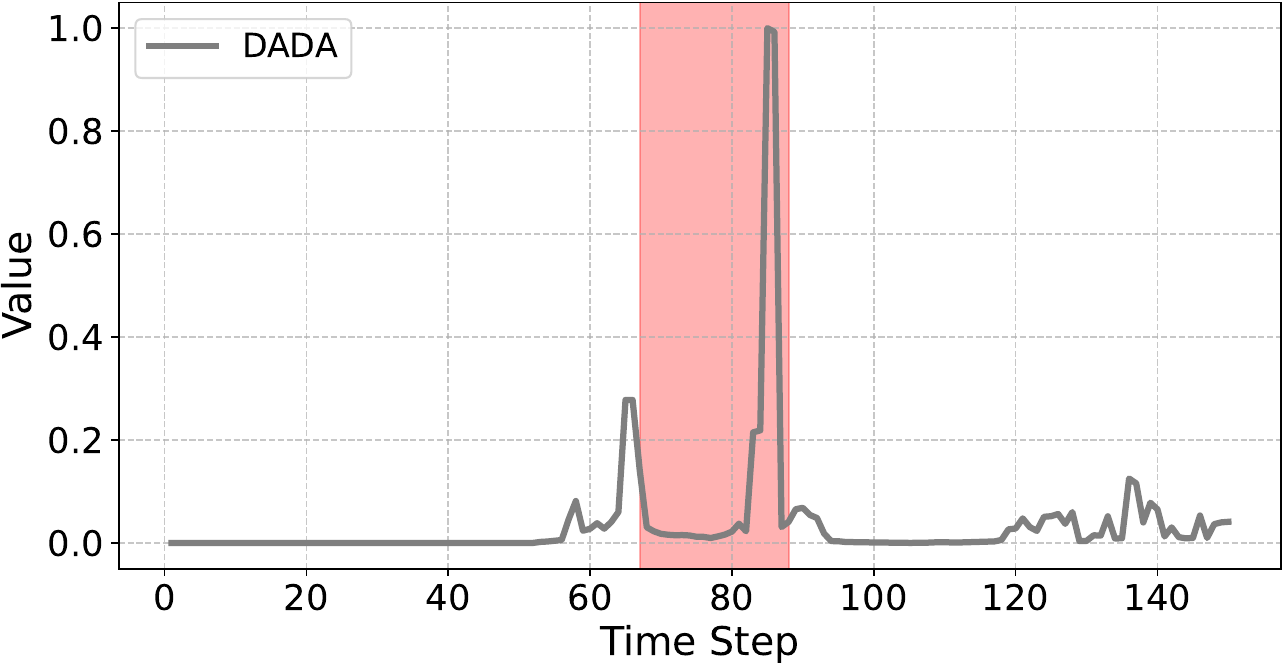}}  
  \hspace{3mm}
  \subfloat[Detect by DADA + STAR]
  {\includegraphics[width=0.3\textwidth]{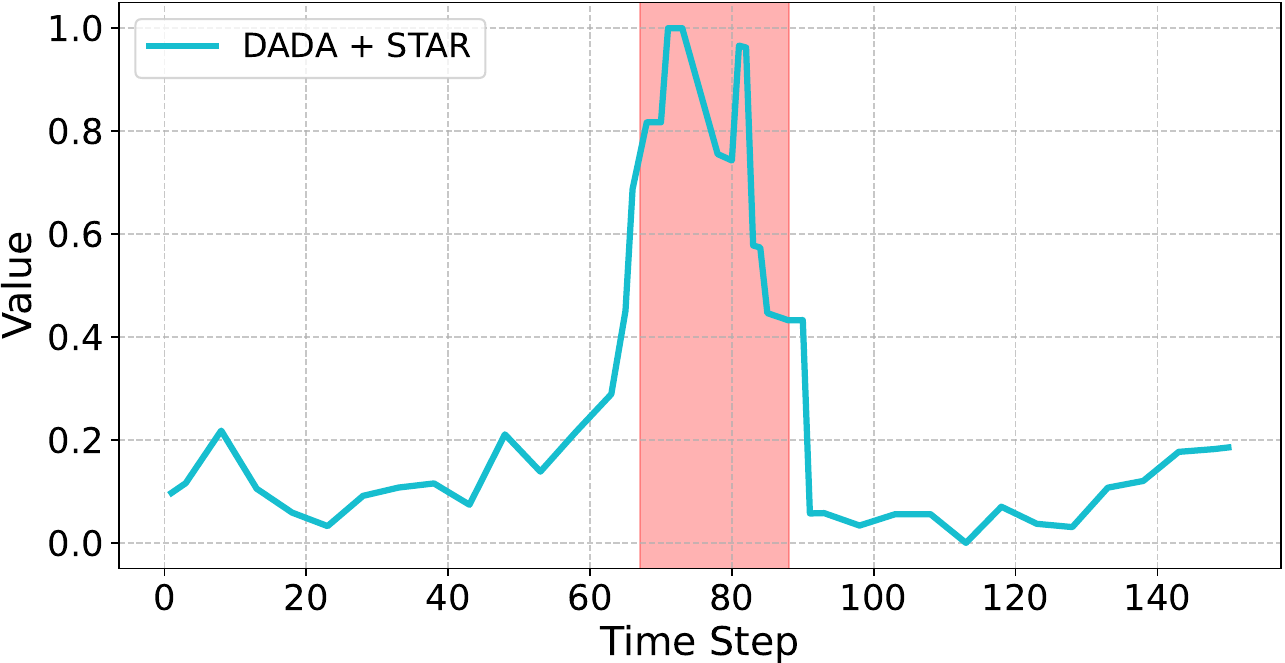}}
    \vspace{3mm}
    \subfloat[Window of case]
  {\includegraphics[width=0.3\textwidth]{Figures/case/case-study-Genesis-data.pdf}}  
  \hspace{3mm}
  \subfloat[Detect by UniTS]
  {\includegraphics[width=0.3\textwidth]{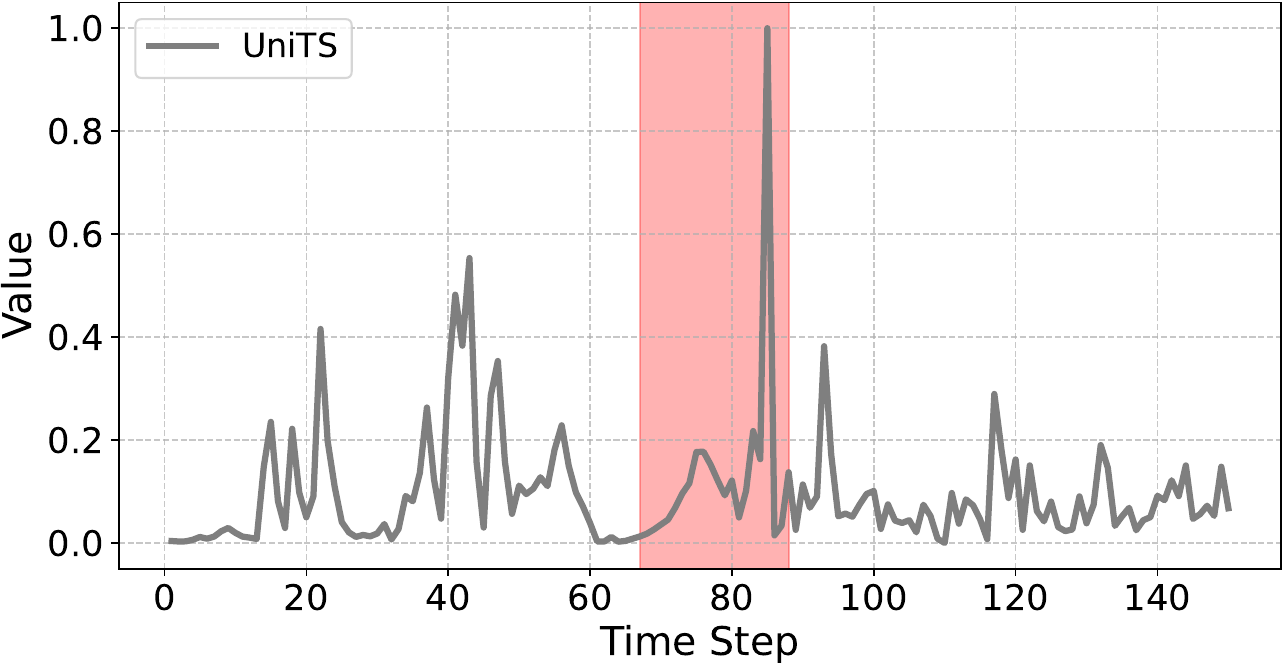}}  
  \hspace{3mm}
  \subfloat[Detect by UniTS + STAR]
  {\includegraphics[width=0.3\textwidth]{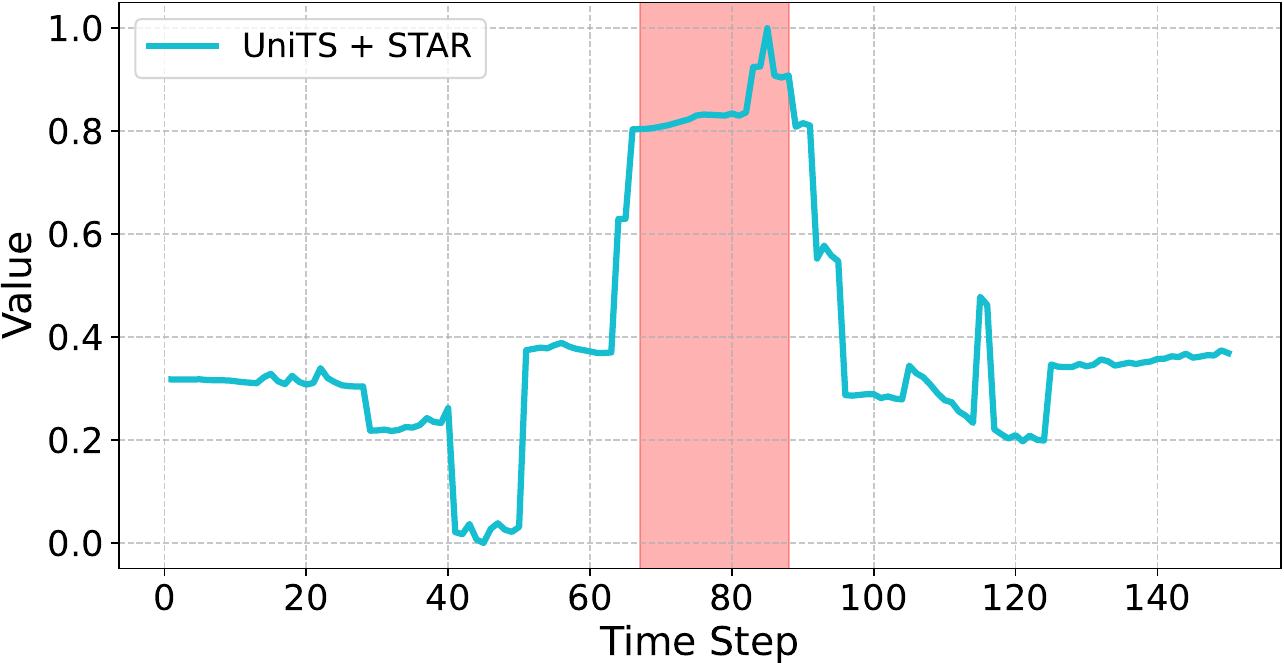}}
 
  \caption{Anomaly score of backbones and STAR in case from Genesis.}
   \label{fig: score genesis}
  \vspace{-3mm}
\end{figure*}

% 为了进一步分析how STAR works, 我们选取了来自不同数据集的case进行了可视化。
To further analyze how STAR works, we selected and visualized cases from real-world datasets.

(1) \textbf{Genesis} Figure~\ref{fig: case Genesis} illustrates that the position state variables (Pos1reached, Pos1reached, Pos1reached, and Pos1reached)  have a condition-based inference on the actual motor speed (ActSpeed). Under normal circumstances, the actual motor speed exhibits a stable pattern when the system is in a state associated with reaching a specific position (PosXreached = 1). However, when the system is not reached at any specific position, the motor speed displays an unstable pattern (PosXreached = 0). 
% 如图5所示，与位置状态有关的变量(Pos1reached)对于电机实际转速(ActSpeed)有着condition-based inference. 在正常情况下，当处于到达某位置的状态时，电机转速处于稳定pattern，但是在不属于任何特定位置时，电机的转速处于不稳定稳定pattern
% case中所显示的异常，则为在not reached at any specific position的状态下，the actual motor speed 却exhibits a stable pattern，出现了state-数值 不匹配的异常。

The anomaly presented in the case is that the actual motor speed exhibits a stable pattern even when the system is in a state of not having reached any specific position. This constitutes a state-value mismatch anomaly. 
% 如图6所示，尽管TSFMs对于该异常最后几个时间点处的突变极其敏感，但是却无法对其它部分完成检测。
As shown in Figure~\ref{fig: score genesis}, although TSFMs are highly sensitive to the abrupt change at the final timestamps of the anomaly, they fail to detect the preceding portion.
% 在加入STAR后，得益于状态感知的设计，对state-value mismatch anomaly变得敏感，因此成果的捕获到该异常。
With the integration of STAR, the model gains heightened sensitivity to the state-value mismatch anomaly and successfully captures this anomaly case.

\begin{figure*}[!t]
    \centering
\includegraphics[width=1\linewidth]{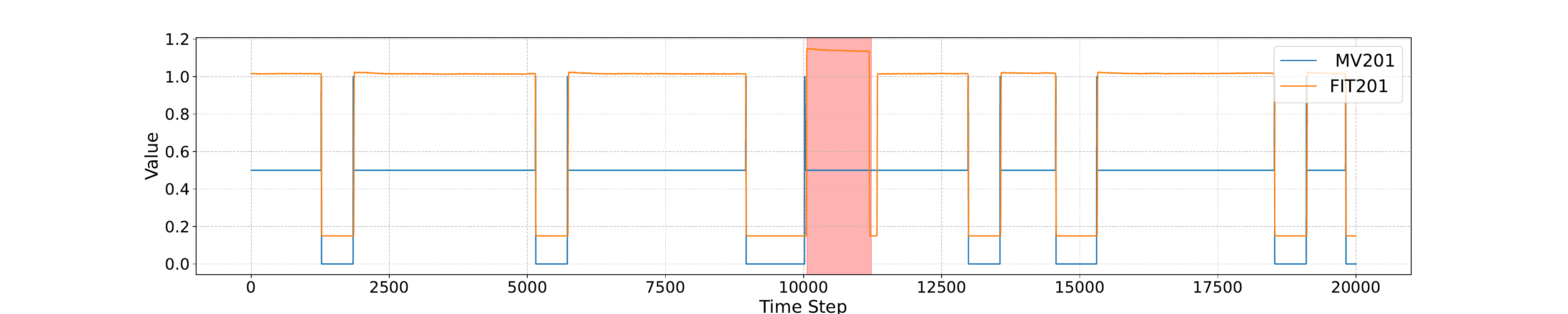}
    \caption{Case series in SWaT.}
\label{fig: case SWAT}
\vspace{-3mm}
\end{figure*}

\begin{figure*}[!t]
  \centering
  \subfloat[Window of case]
  {\includegraphics[width=0.3\textwidth]{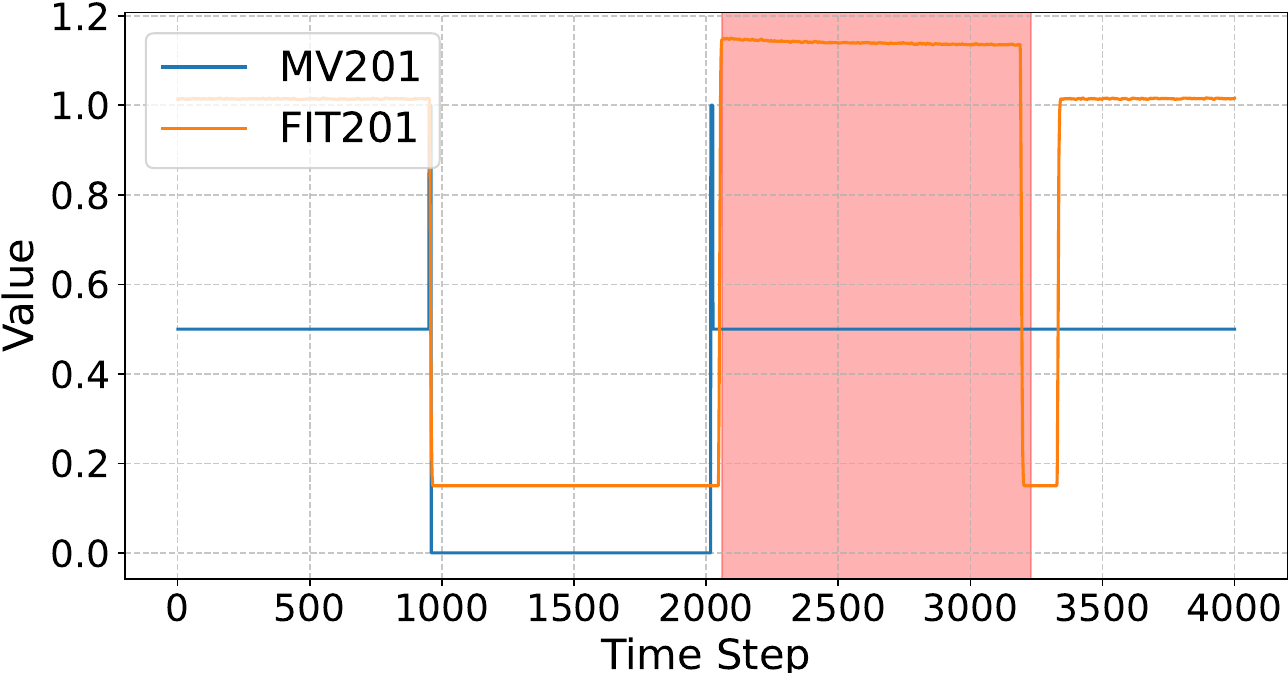}}  
  \hspace{3mm}
  \subfloat[Score of DADA]
  {\includegraphics[width=0.3\textwidth]{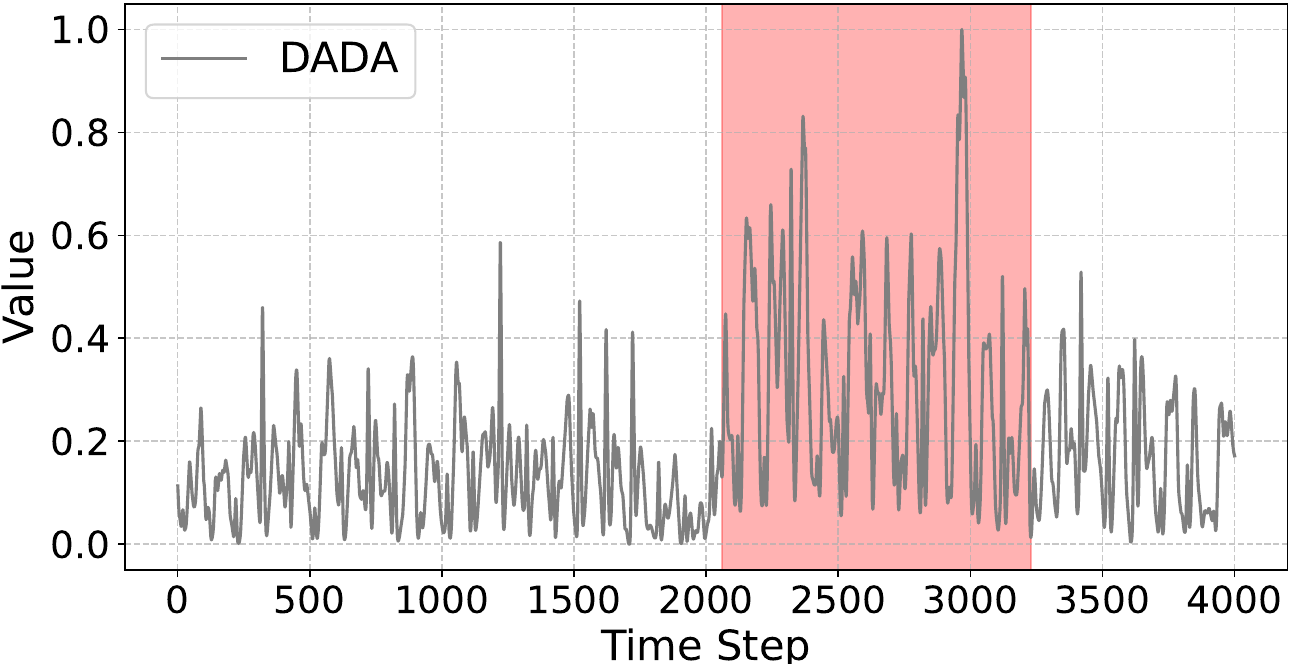}}  
  \hspace{3mm}
  \subfloat[Score of DADA + STAR]
  {\includegraphics[width=0.3\textwidth]{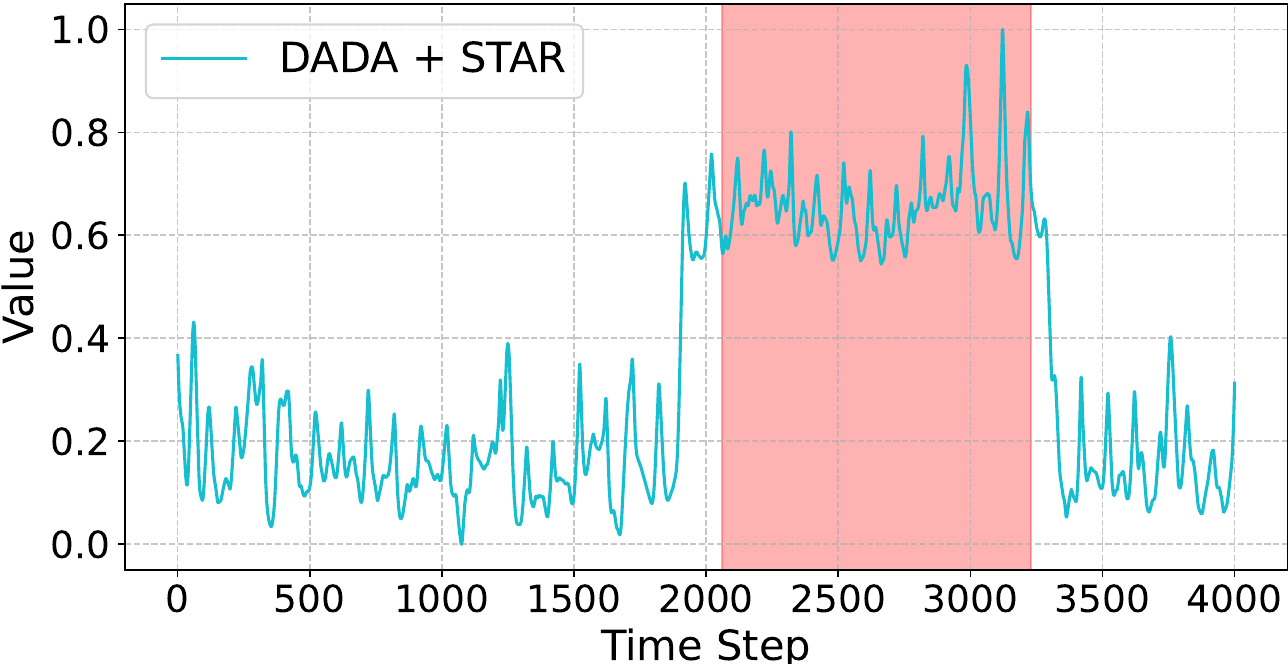}}
    \vspace{3mm}
    \subfloat[Window of case]
  {\includegraphics[width=0.3\textwidth]{Figures/case/case-study-SWAT-data.pdf}}  
  \hspace{3mm}
  \subfloat[Score of UniTS]
  {\includegraphics[width=0.3\textwidth]{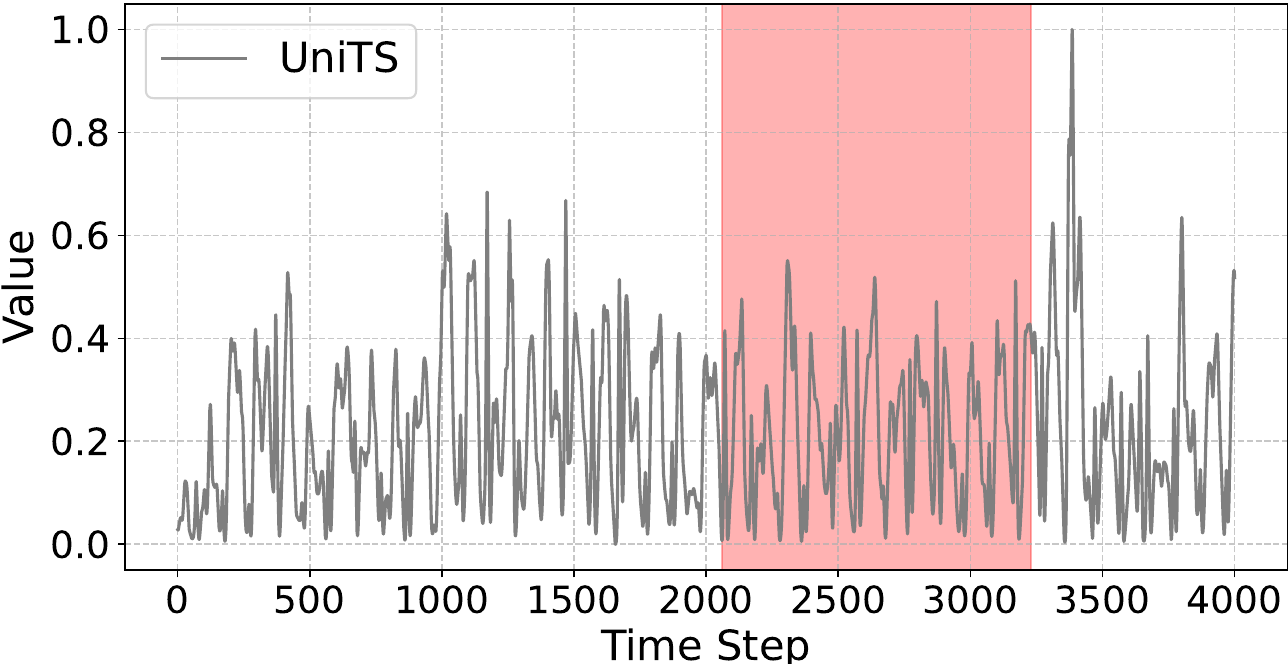}}  
  \hspace{3mm}
  \subfloat[Score of UniTS + STAR]
  {\includegraphics[width=0.3\textwidth]{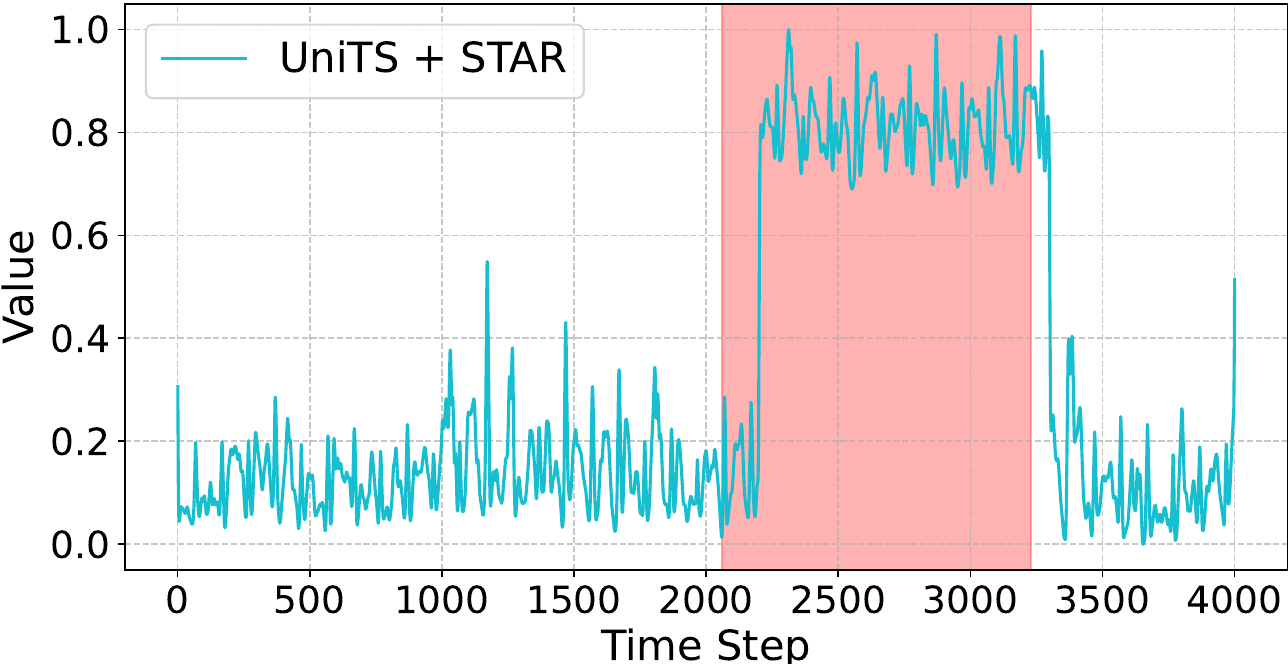}}
  \caption{Anomaly score of backbones and STAR in case from SWaT.}
  \vspace{-3mm}
\end{figure*}

% 如图5所示，与状态变量电动阀门(MV201)对于该区域水流量(FIT201)同样有着condition-based inference. 在正常情况下，当阀门处于打开状态时水流稳定维持较高水平，当阀门处于关闭状态，水流稳定维持较低水平,当阀门处于切换状态，水流快速增长或下降,

(2) \textbf{SWaT} As shown in Figure 5, the state variable of motorized valve switch (MV201) also has a condition-based inference on the water flow (FIT201). Under normal conditions, the behavior is as follows: when the valve is open (MV201=0.5), the water flow remains stable at a high level; when the valve is closed (MV201=0), the flow stabilizes at a low level; and during a transitional state (MV201=1), the flow rapidly increases or decreases.

% case中所显示的异常，一方面表现在水位过高，另一方面表现在未处于transitional state (MV201=1) flow就开始rapidly decreases。
The anomaly presented in the case is twofold. On the one hand, the water level is excessively high. On the other hand, the flow begins to decrease rapidly even though the valve is not in a transitional state (MV201=1), remaining fully open (MV201=0.5).
% 尽管DADA成果的识别出了过高的水位线，但是却对于快速下降的异常却失败了。这可能是由于，该变量确实存在很多短周期内快速下降的模式。
Although DADA successfully identified the excessively high water level, it failed to detect the anomaly of the rapid decrease. This is likely because the variable itself frequently exhibits patterns of rapid decline over short periods.
% STAR通过考虑当前状态与数值的匹配程度，successfully captures this anomaly case.
STAR successfully captures this anomaly by considering the degree of matching between the state and the numerical value.

\section{The Use of Large Language Models}
The use of open-source Large Language Models (LLMs) in this work was strictly limited to assisting with the translation of certain terms and polishing a small portion of the text. LLMs did not contribute to the conceptual aspects of the research, including information retrieval, knowledge discovery, or the ideation process.

\end{document}